\documentclass{article}

\usepackage[final]{corl_2026} 
\usepackage{xcolor}
\usepackage{caption}
\usepackage{xspace}
\usepackage{graphicx}
\usepackage{float}
\usepackage{booktabs}
\usepackage{array}
\usepackage{wrapfig}
\usepackage{longtable}
\usepackage{makecell}
\usepackage{pifont}
\usepackage{amssymb}
\usepackage{amsmath}

\usepackage[normalem]{ulem}

\newcommand{\hmo}{\textsc{HelpMeOut}\xspace}

\title{HelpMeOut: A Neuropsychology-Grounded Benchmark for Error-Aware Monitoring of Household Task Execution}

\author{
  Haoran Wang$^{1}$, \quad Jingwei Huang$^{2}$, \quad Anqing Duan$^{2}$, \quad Ho Seok Ahn$^{1}$, \\
  \quad \textbf{Bruce Macdonald}$^{\dagger, 1}$ \\
  $^{1}$University of Auckland, New Zealand\\
  $^{2}$MBZUAI, United Arab Emirates\\
  $^{\dagger}$\texttt{b.macdonald@auckland.ac.nz}
}

\begin{document}
\maketitle


\begin{abstract}
   We introduce \textsc{HelpMeOut}, an embodied AI benchmark for error-aware monitoring of household task execution. Existing embodied household benchmarks typically assume that humans execute tasks correctly, but this assumption does not hold for cognitively impaired persons or older adults whose everyday activities produce structured execution errors. To address this gap, \textsc{HelpMeOut} operationalizes research on everyday action errors into 20{,}000 paired activity episodes, covering nine neuropsychologically motivated error categories. In each episode, agents observe a streaming human activity trajectory and must monitor whether the execution is unfolding correctly, diagnosing any structured errors as early as possible. We evaluate learned-monitor, symbolic, change-point, and frontier LLM prompting baselines. Results show that these methods perform poorly as error monitors, often missing when execution goes wrong. These results suggest a gap left by standard goal-inference and task-completion benchmarks, making \textsc{HelpMeOut} a testbed for error-aware assistive agents.
\end{abstract}

\keywords{Error-Aware Monitoring, Assistive Robotics, Embodied AI Benchmark} 


\section{Introduction}
\label{sec:intro}
Consider a cognitively impaired person making a cup of instant coffee. She adds the powder and pours hot water, but never stirs---the stirring step does not come to mind. Later, she takes a fork to add sugar, unaware that a spoon is the appropriate tool. These look like minor lapses, but to an assistive robot they pose a distinct problem: the person is not confused about her goal, nor is she asking for help---she is executing an everyday task under degraded cognitive control. How can the robot recognize, from partial household actions, that execution is going wrong? 

Neuropsychological studies of everyday action show that such breakdowns are not random noise: everyday tasks often fail in structured ways  such as omissions, substitutions, sequence errors, and perseverations~\citep{buxbaum1998ideational, giovannetti2002naturalistic}. Detecting such errors and intervening at the right time is a core requirement for assistive robots supporting people with cognitive impairment and older adults. This is a large and growing population: cognitive impairments such as dementia affect tens of millions worldwide---dementia alone reaches 55 million people and is projected to grow to 139 million by 2050~\citep{WHO_Dementia_FactSheet, nichols2022estimation}, and a further 12--18\% of adults over 60 live with mild cognitive impairment, a condition often associated with deficits in instrumental activities of daily living~\citep{olatunji2025robotic}. Yet this capability is rarely evaluated in existing embodied household benchmarks.

\begin{figure}[!t]
  \centering
  \includegraphics[width=\linewidth]{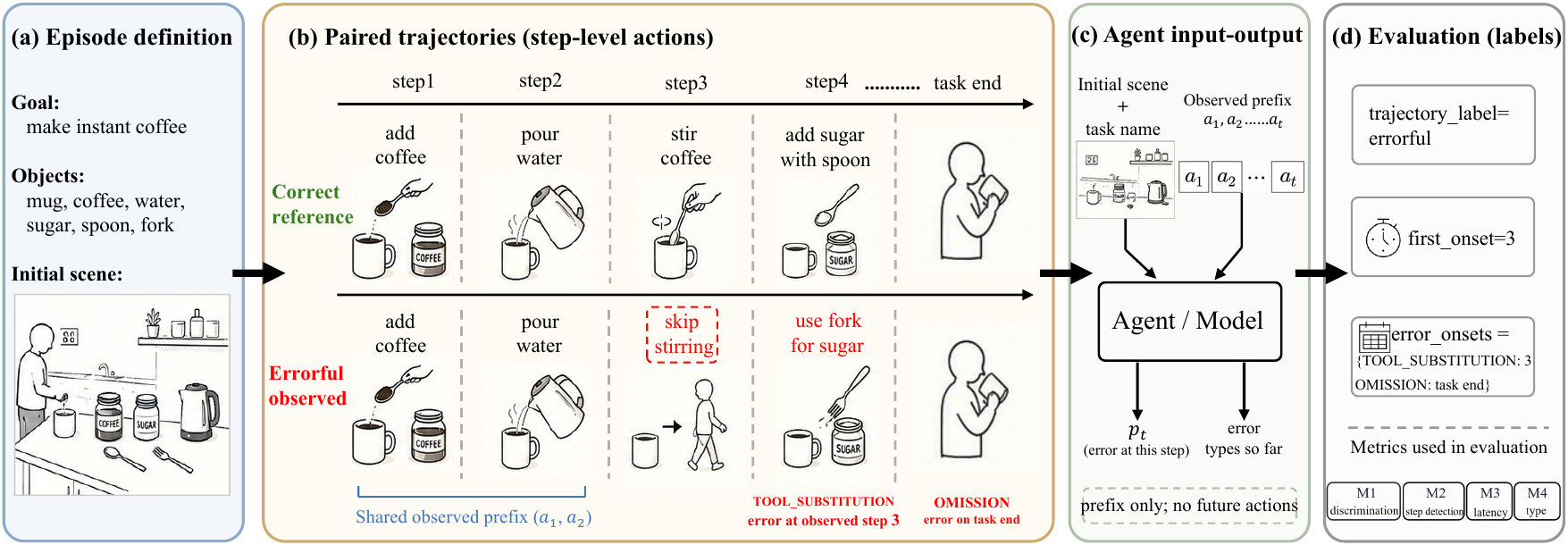}
  \caption{%
  \textsc{HelpMeOut} task overview, shown with an instant-coffee task. \textbf{a.} We define the episode with goal, available objects, and initial scene. \textbf{b. }An episode consists of a pair of trajectories: the correct reference on top and the errorful observed execution on bottom, which contains a \textsc{tool\_substitution} (fork used instead of spoon) and an \textsc{omission} (stirring skipped), with the latter confirmed only when the task ends. \textbf{c. }The system input is the task name, the initial scene, and the action prefix in symbolic PDDL notation; the output is an error probability, onset step, and error type. \textbf{d. }Evaluation labels and the four metrics (M1--M4) derived from the error injection metadata; we explain them in detail in Sec.~\ref{sec:v2-err-val} and Sec.~\ref{sec:v2-eval}.
  }
  \label{fig:story}
  \end{figure}

Most such benchmarks  share a common assumption: humans act coherently toward well-formed goals, and the agent's central problem is inferring \emph{what} the human wants to do. For cognitively impaired persons and older adults, however, the harder question is not only what the person is trying to do, but \emph{whether} their ongoing execution is failing. A separate line of smart-home and robot-assisted systems does target execution failures directly~\citep{das2016one,wilson2019robot,pereyda2019cyber}, but typically relies on fixed sensor setups or predefined activity scripts, yielding task-specific signals rather than standardized, large-scale data for agent evaluation. What is missing is a scalable benchmark that tests step-level error monitoring from partial household activity streams. Figure~\ref{fig:story} illustrates this setup with an instant-coffee episode.

We therefore introduce \textsc{HelpMeOut}, an embodied AI benchmark for error-aware monitoring of household task execution. Guided by the neuropsychological evidence above, \textsc{HelpMeOut} grounds its episodes in a goal-control model~\citep{giovannetti2021goal} whose parameters are calibrated against the per-category error rates reported in naturalistic-action studies of control and clinical populations~\citep{giovannetti2002naturalistic, schwartz2002naturalistic}: error type and frequency vary by error category and degree of cognitive impairment, rather than being sampled uniformly at random. Agents must therefore recognize errors whose structure reflects impaired everyday action, rather than artifacts of the sampling procedure. As in these tests, the task is known; the monitor must judge whether its execution goes wrong.

 \textsc{HelpMeOut} makes three contributions. First, we formalise online error-aware monitoring as a task of deciding whether, when, and what kind of execution error occurs while a household task unfolds. Second, we build a generator that turns Activities of Daily Living~(ADL)/Instrumental Activities of Daily Living~(IADL) tasks into paired correct and errorful episodes, driven by a goal-control model from the neuropsychology of naturalistic action, and release a dataset of 20{,}000 paired episodes\footnote{Code, data, and the symbolic simulator: \url{https://whrcn.github.io/HelpMeOut-Official/}}. Third, we cast the task as a streaming evaluation protocol that separately measures error discrimination, step-level detection, latency, and type diagnosis. Experiments show that learned baselines struggle with step-level error detection and diagnosis, symbolic baselines with additional domain knowledge primarily identify incomplete tasks at episode termination without distinguishing error types, and frontier LLMs remain limited in step-level error localization and timely alerting.

\section{Related Works}

\paragraph{Everyday action errors.}
In the past few decades, clinical neuropsychology studied everyday action impairment using performance-based instruments such as the Naturalistic Action Test (NAT), the Multi-Level Action Test (MLAT), and the Performance Assessment of Self-Care Skills (PASS)~\citep{buxbaum1998ideational,giovannetti2002naturalistic,schwartz1998naturalistic,schwartz2002naturalistic,holm2024performance}. These tests ask participants to perform familiar multi-object tasks and score errors such as omissions, sequence errors, substitutions, additions, and perseverations. The goal-control account~\citep{giovannetti2021goal} draws on the Supervisory Attentional System (SAS) framework~\citep{norman1986attention} 
and contention-scheduling models~\citep{cooper2000contention,cooper2006hierarchical}, linking omissions to weak or decaying task goals and commission errors to failures to suppress competing goal activations, especially around subgoal transitions~\citep{schwartz2006cognitive,schwartz1991quantitative}. An alternative connectionist account models the same error categories 
\begin{table}[!t]
  \centering
  \resizebox{\columnwidth}{!}{%
  \setlength{\tabcolsep}{3pt}
  \begin{tabular}{lcccccccc}
  \toprule
  Benchmark & \makecell{Long-\\horiz.} & \makecell{Object\\interact.} & \makecell{Subgoal\\ctx.} & \makecell{Errorful\\human exec.} & \makecell{Neuropsych err.\\taxonomy} & \makecell{Per-step\\err. labels} & \makecell{Latency\\eval.} & \makecell{Streaming\\eval.} \\
  \midrule
  TextWorld~\citep{cote2018textworld}& \checkmark  & \checkmark & \ding{55} & \ding{55}  & \ding{55}  & \ding{55}  & \ding{55}  & $\triangle$ \\
  ALFRED~\citep{shridhar2020alfred}                           & $\triangle$ & \checkmark & \ding{55}  & \ding{55}  & \ding{55}  & \ding{55}  & \ding{55}  & \ding{55} \\
  VirtualHome~\citep{puig2018virtualhome}                     & \checkmark  & \checkmark & \ding{55}  & \ding{55}  & \ding{55}  & \ding{55}  & \ding{55}  & \ding{55} \\
  Watch-and-Help~\citep{puig2020watch}                        & \checkmark  & \checkmark & \checkmark & \ding{55}  & \ding{55}  & \ding{55}  & \ding{55}  & \ding{55} \\
  BEHAVIOR / 1K~\citep{srivastava2022behavior,li2023behavior} & \checkmark  & \checkmark & \ding{55}  & \ding{55}  & \ding{55}  & \ding{55}  & \ding{55}  & \ding{55} \\
  HandMeThat~\citep{wan2022handmethat}                        & \checkmark  & \checkmark & \checkmark & \ding{55}  & \ding{55}  & \ding{55}  & \ding{55}  & \ding{55} \\
  BALI~\citep{ghose2026open}             & \checkmark  & \checkmark & \checkmark & \ding{55}  & \ding{55}  & \ding{55}  & \ding{55}  & \ding{55} \\
  \midrule
  \textbf{\textsc{HelpMeOut} (ours)}                          & \checkmark  & \checkmark & \checkmark & \checkmark & \checkmark & \checkmark & \checkmark  & \checkmark \\
  \bottomrule
  \end{tabular}
  }
  \caption{Comparison with related benchmarks. \checkmark~present; \ding{55}~absent; $\triangle$~partial. Existing benchmarks evaluate goal inference or task completion under coherent human execution. \textsc{HelpMeOut} introduces errorful execution, a neuropsychological error taxonomy, step-level error labels, detection-latency evaluation, and streaming evaluation.}
  \label{tab:benchmarks}
  \end{table}
through recurrent dynamics without explicit schema hierarchies~\citep{botvinick2004doing}. More recently, these error taxonomies have been extended to characterise subtle, sub-clinical errors in mild cognitive impairment~\citep{seligman2014subtle}.

\paragraph{Assistive error detection systems.}
Smart-home and robot-assisted systems have used sensors or predefined activity scripts to detect deviations and prompting opportunities~\citep{das2016one,wilson2019robot,pereyda2019cyber}. Deployed prompting systems include COACH, which uses a POMDP over video observations to guide dementia patients~\citep{mihailidis2008coach,hoey2010automated}, and SmartPrompt2, which targets common omission errors~\citep{hackett2026smartprompt}. Related datasets capture erroneous kitchen and procedural activities: CaptainCook4D records egocentric cooking videos in which participants deviate from recipes to induce errors~\citep{peddi2024captaincook4d}, while the Kitchen Task Assessment dataset records sensor data from participants simulating typical dementia errors~\citep{yordanova2020kitchen}. Existing methods also address procedural error detection: PREGO compares recognized steps against LLM-predicted steps~\citep{flaborea2024prego}, differentiable task-graph learning checks steps against a learned task graph~\citep{seminara2024dtgl}, and MistExit aims to identify erroneous pre-segmented key steps as early as possible~\citep{majumder2026mistexit}. These works demonstrate the importance of detecting execution failures, but are often constrained by specific tasks, sensor setups, or errors that participants induce or simulate. As a result, they do not provide a scalable benchmark for evaluating whether models can detect neuropsychologically grounded error types as they unfold, with explicit onset labels and first-alarm timing metrics. Large-scale data collection from cognitively impaired users is also constrained by privacy and ethical concerns. \textsc{HelpMeOut} addresses these limitations by converting the neuropsychological taxonomy and goal-control account into automatically labelled episodes, with clinically calibrated error rates and deterministic step-level onsets, for streaming agent evaluation.

\paragraph{Embodied household benchmarks and simulators.}
Monitoring an ongoing task is closely related to goal and plan recognition from partial observations~\citep{ramirez2010probabilistic,baker2009action}. Classical methods usually assume that the observed agent acts rationally; bounded-rational variants relax this assumption by allowing sub-optimal plans~\citep{zhixuan2020online}, which is closer to the impaired executions considered in \textsc{HelpMeOut}; such work infers goals from imperfect behavior, whereas \textsc{HelpMeOut} asks whether, when, and how the execution itself goes wrong. LLMs are also increasingly used as embodied household planners~\citep{ahn2022saycan,huang2022inner}, motivating their evaluation as error monitors in our benchmark.

Two adjacent lines of work share \textsc{HelpMeOut}'s household setting. The first evaluates agents that infer human intent or resolve instructions from observed behavior and language, including HandMeThat~\citep{wan2022handmethat}, FISER~\citep{wan2025infer}, BALI~\citep{ghose2026open}, and Watch-and-Help~\citep{puig2020watch}. The second provides scalable household task simulators and trajectory generators, including TextWorld~\citep{cote2018textworld}, BEHAVIOR and BEHAVIOR-1K~\citep{srivastava2022behavior,li2023behavior}, VirtualHome~\citep{puig2018virtualhome}, ALFRED~\citep{shridhar2020alfred}, and Habitat rearrangement~\citep{szot2021habitat}. Across both lines, however, the human action stream is assumed competent and the resulting trajectories are correct by construction. \textsc{HelpMeOut} keeps the scalable household-generation setting but adds neuropsychologically grounded errorful trajectories, paired correct references, and step-level onset labels for error-aware monitoring (Table~\ref{tab:benchmarks}).

\section{The \textsc{HelpMeOut} Benchmark}

\subsection{Preliminaries: Neuropsychology of Everyday Action}
\label{sec:v2-prelim}
\textsc{HelpMeOut} uses the neuropsychological literature on everyday action impairment to define both its error labels and its generation model. Performance-based instruments such as NAT, MLAT, and PASS identify recurring failures in familiar multi-object tasks, including omissions, sequence errors, substitutions, additions, and perseverations~\citep{buxbaum1998ideational,giovannetti2002naturalistic,schwartz1998naturalistic,schwartz2002naturalistic,holm2024performance}. These categories recur across clinical populations such as CVA, traumatic brain injury, and dementia, and provide the basis for our benchmark labels.

\begin{figure}[!t]
  \centering
  \includegraphics[width=\linewidth]{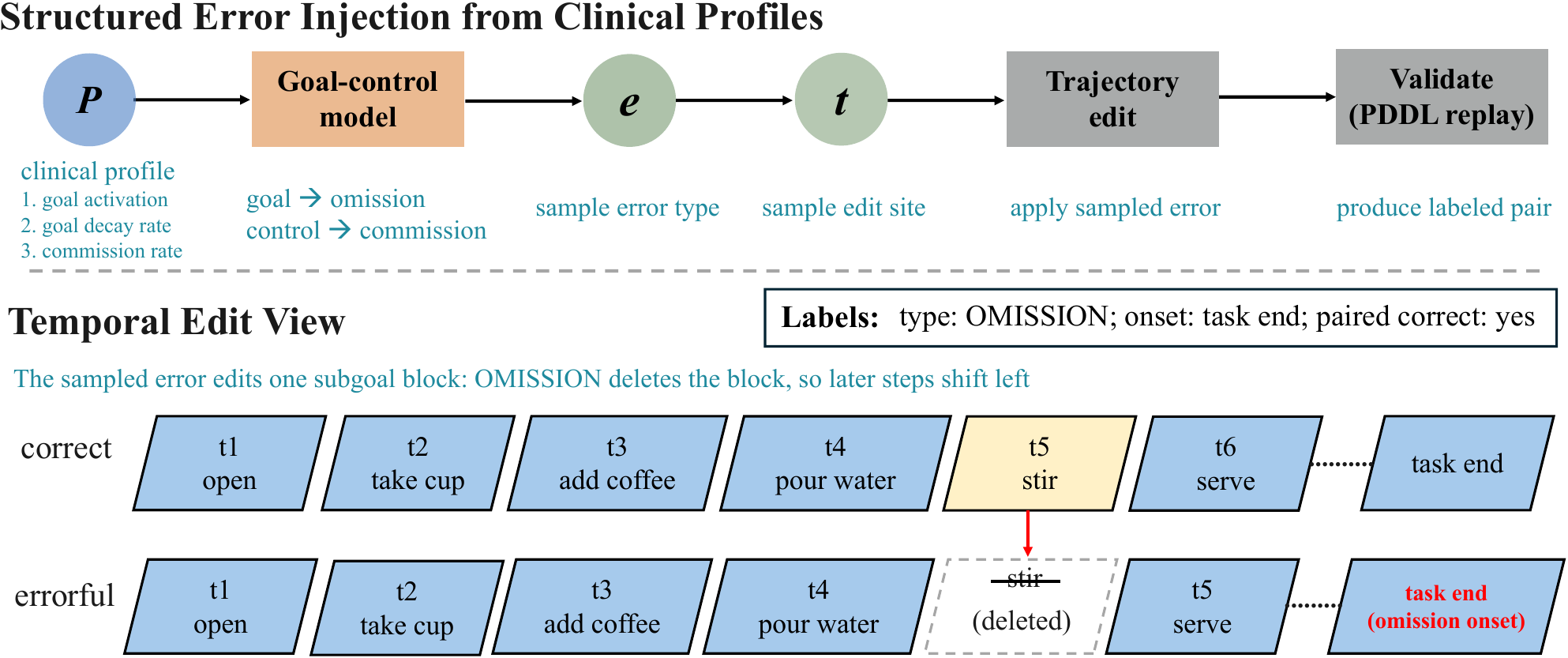}
  \caption{%
  Neuropsychology-based error-generation pipeline. A neuropsychological profile drives the goal-control model to sample an error type and the subgoal block to edit; the edited trajectory is validated by symbolic replay, and onsets are derived by category-specific rules. The temporal view shows an \textsc{omission} deleting one subgoal block and shifting later steps left; the omission is confirmed at the task-end event.%
  }
  \label{fig:benchmark-framework}
  \end{figure}

For generation, we adopt the goal-control account~\citep{giovannetti2021goal}, which separates failures of goal activation from failures of executive control. In our benchmark, goal-activation failures produce omissions, while executive-control failures produce commission errors such as substitutions, additions, and captures. This abstraction gives the generator two interpretable control factors: whether a required subgoal is omitted, and whether a competing action is inserted or substituted. Sec.~\ref{sec:v2-err-val} describes how these factors are instantiated in the error generator. Figure~\ref{fig:benchmark-framework} summarizes this generation process. Additional background on the neuropsychological literature is provided in the supplementary material.

\subsection{Basic Monitoring Setting}
\label{sec:v2-formal}

\textsc{HelpMeOut} is built over a symbolic household domain with 156 object types and 38 grounded action operators covering object manipulation, tool use, and state transformations. At evaluation step \(t\), an agent observes the task name, the initial symbolic scene representation, and the action prefix so far, including the current action. The task is known to the monitor, as in the naturalistic-action tests on which our error categories are based: in the NAT and MLAT, the examiner names the task and its requirements before the person begins, and errors are scored against those requirements~\citep{schwartz1998naturalistic,schwartz2002naturalistic}. Unlike the examiner, the monitor receives only the task name, not its formal requirements, and must infer what the task calls for. From this prefix, the agent must answer three questions: whether execution has already gone wrong, when the observable error onset occurred, and which error categories have appeared.

Each \textsc{HelpMeOut} episode consists of paired trajectories. A symbolic planner first generates a correct household execution from an ADL/IADL task template, and the error process then modifies that execution to obtain an errorful counterpart. The templates cover familiar multi-object routines such as food preparation, packing, cleaning, laundry, medication, and personal routines, following task families used in clinical and functional assessments of everyday action~\citep{schwartz2002naturalistic,holm2024performance,lawton1969assessment,baum2008reliability,katz1963studies}. Formally, we write each episode as
\begin{equation}
e=\langle s_0, \Omega_{\mathrm{correct}}, \Omega_{\mathrm{err}},
\mathcal{S}_0,\ y_{0:T'-1},\ O \rangle .
\end{equation}

Here, \(s_0\) is the initial household state, \(\Omega_{\mathrm{correct}}\) is the reference execution, and \(\Omega_{\mathrm{err}}\) is the realized errorful execution of length \(T'\). \(\mathcal{S}_0\) is the initial symbolic scene representation exposed to agents, encoding object identities, locations, states, affordances, and spatial relations. The sequence \(y_{0:T'-1}\) gives sparse onset annotations: a step is labelled with \texttt{has\_error}\(=1\) only when an injected error first becomes observable. \(O=\{(\mathrm{type}_j,t_j)\}_j\) records each error category and onset step, which supports step-level error detection, first-alarm timing, and cumulative error-type diagnosis. Subgoal boundaries and generation goals are retained as dataset metadata for constructing and analysing episodes; they are not shown to agents. Every episode ends with an explicit task-end event, at which the person declares the whole task finished.

An error's observable onset is the earliest step at which it can be identified from the observed execution prefix and task constraints, using category-specific rules: the first use of a substitute object or tool, an out-of-order action, an action repetition, committing an unprocessed item, placing an unrelated item in the task container, acting on an unrelated object, or leaving an unrelated storage location empty-handed. An omission removes an entire subgoal block. Since valid subgoal orders vary, a missing subgoal may still be completed later; thus, omission becomes observable only at the task-end event. An earlier consequence, such as storing an unprocessed item, is instead labelled as anticipation under MLAT coding~\citep{schwartz1998naturalistic}. Computed automatically from $s_0$, $\Omega_{\text{err}}$, and the injected category, onset annotations are deterministic and reproducible without human raters.

\subsection{Structured Execution Errors}
\label{sec:v2-err-val}
\paragraph{Goal-Control error process.}
Inspired by the goal-control account~\citep{giovannetti2021goal}, we treat omissions as goal-activation failures and commission errors as executive-control failures. We operationalise this distinction by computing, for each subgoal block \(b_k\), an activation level:
\begin{equation}
u_k = \theta_{\mathrm{goal}}(1-\theta_{\mathrm{decay}})^k + \epsilon_k, \qquad \epsilon_k \sim \mathcal{N}(0,\sigma^2),
\end{equation}
where \(\theta_{\mathrm{goal}}\) controls baseline goal salience and \(\theta_{\mathrm{decay}}\) captures accumulating cognitive load across subgoals. If \(u_k < \tau_{\mathrm{omit}}\), the subgoal is omitted; otherwise, with probability \(\theta_{\mathrm{ctrl}}\), a commission error is sampled from a task-conditioned distribution modulated by per-category feasibility. We calibrate $\theta_{\mathrm{goal}}, \theta_{\mathrm{decay}}, \theta_{\mathrm{ctrl}}$ and the category weights $w_c$ so that the marginal error distribution produced by the generator matches the per-category error rates reported for control and clinical populations by \citet{schwartz2002naturalistic} and \citet{giovannetti2002naturalistic}. The resulting impaired profile ($\theta_{\mathrm{goal}}{=}0.48$, $\theta_{\mathrm{decay}}{=}0.20$, $\theta_{\mathrm{ctrl}}{=}0.12$) aggregates rates across left-CVA, right-CVA, TBI, and dementia subjects in those two studies. Numerical values of $\tau_{\mathrm{omit}}$ and the per-category feasibility scores $\phi_c(s_k)$ are reported in the supplementary material.

\paragraph{Error taxonomy.}
The nine error categories are grounded in MLAT/NAT taxonomies summarized by the goal-control model~\citep{giovannetti2021goal}. We organise them into two levels: progression-level errors that disrupt the structure of task execution (\textsc{omission}, \textsc{sequence}, \textsc{anticipation}, \textsc{perseveration}), and current-action errors that primarily affect local action selection, object use, or spatial targeting (\textsc{object\_substitution}, \textsc{tool\_substitution}, \textsc{spatial\_misreach}, \textsc{action\_addition}, \textsc{capture}). Detailed definitions and examples for each error category are provided in the supplementary material. This distinction matters for assistance: progression-level errors require reactivation or redirection of the task plan, whereas current-action errors require immediate correction of the ongoing action. Consistent with clinical reports across neurological populations, omissions are treated as a prominent error class, while perseverations are sampled sparsely to reflect their lower base rate. Correct executions vary in subgoal order and intermediate placements, as the generator randomizes subgoal order and plans each subgoal separately. A reordering is labelled \textsc{sequence} only if it violates a required order (stirring before all ingredients are added, or a step-order norm such as buttering bread before toasting); otherwise, it remains correct.

\paragraph{Dual-standard validation.}
A key challenge is that errorful trajectories must be semantically wrong yet physically executable: a naive PDDL validator would reject substitution errors, because using a fork where a spoon is required is the violation, while an unconstrained validator admits physically impossible traces. We resolve this by applying strict precondition replay to the seven non-substitution error types, and for substitution errors enforcing only the physical chain---navigation, accessibility, containment, hand occupancy---while relaxing the single semantic precondition that defines the substitution.
This ensures that injected errors remain interpretable as human execution errors rather than simulator failures: every errorful trajectory is physically executable, and the person can carry out each of its actions. 

\paragraph{Dataset statistics.}
The \textsc{HelpMeOut} benchmark spans 50 ADL/IADL task templates and contains 20,000 episodes---12,000 errorful and 8,000 correct, with correct episodes paired with themselves and labeled error-free. Errorful trajectories average 24.7 steps, including the task-end event, with 1.07 injected errors each (6.8\% have more than one), yielding 12{,}746 annotated onsets; omissions are the most frequent category (5{,}346), followed by sequence errors (1{,}478) and object substitutions (1{,}142). We hold out 500 test episodes (312 errorful, 188 correct) and a 2{,}022-episode validation split; the remaining 17{,}478 episodes are for training.

\subsection{Prefix-Based Evaluation}
\label{sec:v2-eval}
We evaluate online error monitoring as a streaming task: at step $t$, the agent observes the task name, $\mathcal{S}_0$, and the action prefix $a_{0:t}$, and outputs $p_t$, the probability that an error occurs at step $t$, together with cumulative probabilities for the nine categories; no future action is visible. For a trajectory with $N$ actions, the first $\lfloor N/4 \rfloor$ steps are a context window that is not scored, and all onsets fall in the remaining evaluation window, which ends with the task-end event. The four metrics correspond to the benchmark questions of whether something went wrong, where the error becomes observable, when the model raises its first alarm, and what type of error occurred:

\textbf{M1---Trajectory-level discrimination.} The score of a trajectory is $s=\max_t p_t$ over the evaluation window, and AUROC measures whether errorful trajectories score above correct ones. M1 does not depend on when the alarm is raised; timeliness is measured by M2 and M3.
\textbf{M2---Step-level detection.} Each evaluated step is scored against its label $y_t$ with AUPRC. Only an onset step is positive; later steps, where the error's consequences persist, are negative, so a late alarm earns no credit and counts as a false alarm. Because omissions are confirmed at the task-end step, we also report M2-end, the AUPRC at that step, and M2-mid, the AUPRC at the earlier steps where all other errors occur.
\textbf{M3---First-alarm timing.} For the first-alarm step $\hat{t}=\min\{t:p_t\ge\theta\}$ with $\theta=0.5$, latency is $\Delta=\hat{t}-t_{\mathrm{onset}}$. We group alarms as early ($\Delta<0$), on target ($\Delta=0$), late ($\Delta>0$), or missed, and report the false-alarm rate on correct episodes. Omissions, which can only be confirmed at task end, are reported separately.
\textbf{M4---Error-type diagnosis.} The cumulative category probabilities are scored against the set of categories that have occurred so far (multi-label): in errorful trajectories from the first onset onward, and in correct trajectories at every evaluated step, where this set is empty and naming any category is a false positive. Each trajectory is weighted equally, and we report macro-F1. A monitor is thus credited for naming an error's category only where an error has actually occurred.

\begin{table}[t]
\centering
\small
\setlength{\tabcolsep}{4.5pt}
\renewcommand{\arraystretch}{1.08}
\caption{Main results on the 500-episode test set (metrics in Sec.~\ref{sec:v2-eval}). M3 at $\theta{=}0.5$ over the 175 episodes with an error before task end; FA over the 188 correct episodes (re-planner: timed-out episodes excluded). Random: positive-step rate for M1 and M2, a uniformly random monitor for M3 and M4 (mean of 20 seeds). $^\dagger$Human reference on the 60-episode subset (M2 random 0.027; M3 over 26 errorful and 30 correct episodes). Symbolic references use extra knowledge (Sec.~\ref{sec:result}).}
\label{tab:main}
\begin{tabular}{@{}lcccccccc c@{}}
\toprule
 & & \multicolumn{3}{c}{M2} & \multicolumn{4}{c}{M3} & \\
\cmidrule(lr){3-5}\cmidrule(lr){6-9}
Method & M1 & all & end & mid & Missed & Early & On target & FA & M4 \\
\midrule
Random            & 0.500 & 0.036 & 0.274 & 0.023 & 0.00 & 0.89 & 0.05 & 1.00 & 0.116 \\
Human$^\dagger$   & 0.967 & 0.748 & --    & --    & 0.12 & 0.15 & 0.73 & 0.17 & 0.778 \\
\midrule
\multicolumn{10}{@{}l}{\emph{Learned and change-point monitors}} \\
Seq2Seq           & 0.496 & 0.376 & 0.656 & 0.132 & 0.99 & 0.00 & 0.00 & 0.10 & 0.502 \\
Seq2Seq+SG        & 0.554 & 0.394 & 0.665 & 0.151 & 0.98 & 0.00 & 0.00 & 0.05 & 0.548 \\
SGT               & 0.778 & 0.656 & 0.791 & 0.553 & 0.52 & 0.04 & 0.38 & 0.12 & 0.542 \\
CUSUM             & 0.645 & 0.039 & 0.235 & 0.029 & 0.06 & 0.77 & 0.07 & 0.81 & 0.000 \\
\midrule
\multicolumn{10}{@{}l}{\emph{Symbolic references (extra knowledge)}} \\
Precondition check & 0.513 & 0.059 & 0.274 & 0.062 & 0.95 & 0.00 & 0.05 & 0.00 & 0.034 \\
Re-planner         & 0.887 & 0.379 & 0.690 & 0.118 & 0.10 & 0.75 & 0.05 & 0.62 & 0.138 \\
\bottomrule
\end{tabular}

\vspace{8pt}
\caption{SGT by error category on the test set. Each category is scored against all correct episodes: M1 over episodes, M2 over steps (for omissions, all evaluated steps with the task-end step as the positive; for other categories, the steps before task end), with Rand.\ its positive-step rate. On target and missed ($\theta{=}0.5$) are shares of the category's episodes; M4 is per-class F1.}
\label{tab:sgt_class}
\begin{tabular}{@{}lrcccccc@{}}
\toprule
Category & $n$ & M1 & M2 & Rand. & On target & Missed & M4 F1 \\
\midrule
Omission             & 137 & 0.831 & 0.730 & 0.026 & 58\% & 41\% & 0.916 \\
Sequence             & 38  & 0.891 & 0.814 & 0.010 & 74\% & 26\% & 0.921 \\
Anticipation         & 17  & 0.861 & 0.359 & 0.005 & 29\% & 47\% & 0.607 \\
Perseveration        & 26  & 0.725 & 0.502 & 0.007 & 42\% & 54\% & 0.803 \\
Object substitution  & 30  & 0.522 & 0.038 & 0.008 & 0\%  & 87\% & 0.196 \\
Tool substitution    & 9   & 0.501 & 0.020 & 0.003 & 0\%  & 89\% & 0.000 \\
Action addition      & 9   & 0.405 & 0.029 & 0.003 & 11\% & 89\% & 0.216 \\
Capture              & 23  & 0.686 & 0.178 & 0.012 & 13\% & 70\% & 0.502 \\
Spatial misreach     & 23  & 0.952 & 0.771 & 0.007 & 83\% & 4\%  & 0.715 \\
\bottomrule
\end{tabular}
\end{table}

\section{Experimental Results}
\label{sec:result}

\paragraph{Evaluation protocol.}

We evaluate three families of monitors. \emph{Learned}: Seq2Seq, which encodes the initial scene and the observed action prefix; Seq2Seq+SG, which additionally encodes the symbolic scene graph replayed from the PDDL state (object identity, location, state, and containment; no geometry); and a scene-graph transformer (SGT) that reads the replayed graph before each action together with the action itself. \emph{Non-learned}: a CUSUM change-point detector over action-bigram surprise fitted on correct training trajectories; a precondition checker that flags actions whose PDDL preconditions fail; and a re-planner that, given the task's template-level goal, plans with the generator's planner, tracks the plan, re-plans when the observed action deviates, and raises an alarm when the re-plan grows longer, when an action cannot execute, or when the goal is unmet at task end. \emph{LLMs}: frontier models prompted step by step with the same prefix-only input. All learned and non-learned baselines run on the full 500-episode test set; because LLM evaluation is costly, LLMs and the Human reference use a 60-episode subset (30 errorful, 30 correct). For the Human reference, annotators saw the task setting (objects, initial scene, and error categories) and judged each prefix under the same whether, when, and what-kind questions. We report 95\% confidence intervals from a trajectory-level bootstrap stratified by errorful and correct episodes, and we do not rank methods whose intervals overlap. Implementation details are in the supplementary material.

\begin{figure}[!t]
\centering
\includegraphics[width=1\linewidth]{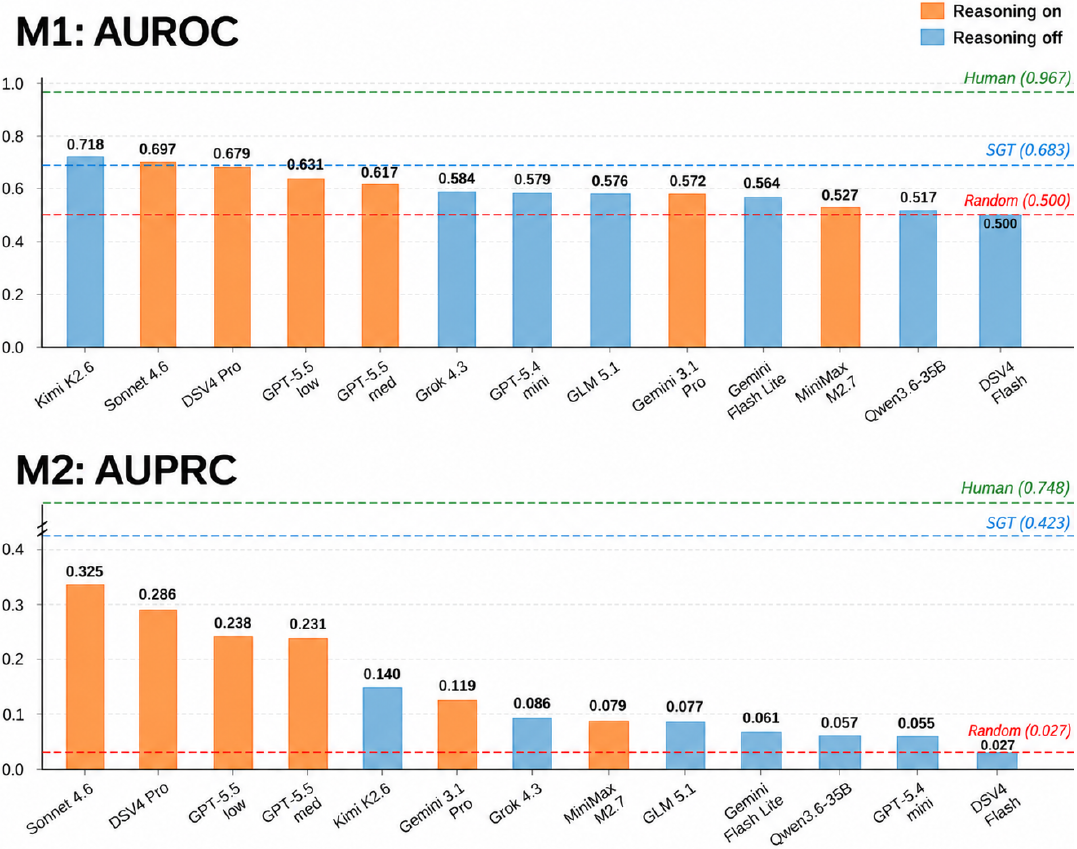}

\caption{M1 AUROC and M2 AUPRC of the thirteen LLM configurations on the 60-episode subset; dashed lines mark the Human, SGT, and Random references, and color marks whether reasoning is enabled. Confidence intervals and full M1--M4 results are in the supplementary material.}
\label{fig:llm_overview}
\end{figure}
\paragraph{Learned and change-point baselines.}Table~\ref{tab:main} summarizes the results. Seq2Seq and Seq2Seq+SG detect error steps well above chance on M2 (0.376 and 0.394 vs.\ 0.036), but most of this comes from the task-end step, where omissions are confirmed (end-step AUPRC 0.66 vs.\ 0.13--0.15 at the remaining steps), and their trajectory-level discrimination stays near chance (M1 0.496 and 0.554). Seq2Seq+SG improves over Seq2Seq on M1 ($+$0.058, 95\% CI $[+0.023, +0.092]$) and M4 ($+$0.047, $[+0.019, +0.077]$), but not significantly on M2. SGT, which conditions every step on the current action, is the strongest learned monitor (M1 0.778, M2 0.656; 0.553 at non-final steps); its gains concentrate on errors with local action signatures (sequence, spatial misreach, perseveration), while substitutions and action additions remain near chance (Table~\ref{tab:sgt_class}). CUSUM reaches M1 0.645 only by alarming on 81\% of correct episodes, with M2 at chance and no type diagnosis, which is why M1 alone cannot measure monitoring. All methods remain far below Human (M1 0.967, M2 0.748).

\paragraph{Symbolic baselines.}
Unlike the learned monitors, the symbolic baselines use the PDDL domain model, and the re-planner additionally receives the task's formal goal and the generator's planner; they are knowledge-rich references rather than competitors, and they test whether the benchmark's errors can be caught by construction. Because seven of the nine categories are executable under strict PDDL replay, the precondition checker fires on fewer than 5\% of errorful episodes (M1 0.513). The re-planner attains the highest M1 (0.887), but almost entirely through its end-of-task check: once the person declares the task finished, an unmet goal confirms an omission with no false alarms, and this check alone gives M1 0.802. Restricted to steps before task end, its M1 drops to 0.669 $[0.622, 0.714]$, overlapping with the other baselines (0.59--0.62); its step-level AUPRC at non-final steps is 0.118 (SGT: 0.553); and 62\% of correct episodes receive an alarm, because valid human executions are not shortest plans and deviation from a plan is not an error. A symbolic monitor can thus confirm after the fact that a task was left incomplete, but it does not detect or localize errors while execution unfolds.

\paragraph{Alarm timing and type diagnosis.}
At $\theta{=}0.5$, the Seq2Seq models almost never raise an alarm (miss rate 0.99 and 0.98): their per-step scores rank error steps above other steps but rarely cross the threshold. SGT is the only monitor whose alarms are mostly timely---80\% of its first alarms fall at the onset step, at a 12\% false-alarm rate on correct episodes---but it misses 52\% of errorful episodes, so only 38\% receive an on-target alarm; the misses concentrate in the substitution, action-addition, and capture categories (70--89\% missed), whereas 83\% of spatial misreaches receive an on-target alarm (Table~\ref{tab:sgt_class}). CUSUM and the re-planner rarely miss, but their first alarm comes early in about three quarters of errorful episodes, and they alarm on 81\% and 62\% of correct episodes. On M4, the learned monitors reach macro-F1 0.50--0.55 (per-class F1 in Table~\ref{tab:sgt_class} and the supplementary material). Their type heads also name an error category in most correct episodes (all 188 for both Seq2Seq models, 85\% for SGT), which M4 penalizes; scored on errorful episodes alone, their macro-F1 would be 0.63--0.68. The Seq2Seq models thus name error categories confidently while almost never raising an alarm, so their type and alarm outputs are not consistent with each other.

\paragraph{LLM baselines do not close the monitoring gap.}Figure~\ref{fig:llm_overview} compares thirteen LLM configurations on the 60-episode subset. The upper ends of the best LLM confidence intervals stay below Human on both M1 (0.837 vs.\ 0.967) and M2 (0.482 vs.\ 0.748), and on M2 no LLM exceeds SGT on the same episodes (0.423). Intervals overlap substantially across models, so we do not rank them. Reasoning matters most for localization and diagnosis: with reasoning enabled, the larger models (Claude Sonnet 4.6, DeepSeek V4 Pro, GPT-5.5, Gemini 3.1 Pro) reach M2 0.12--0.33 and M4 0.43--0.52, whereas with reasoning disabled M2 stays at or below 0.14 and M4 at or below 0.31. More reasoning effort does not help further (GPT-5.5 low vs.\ medium: M2 0.238 vs.\ 0.231). With reasoning enabled, these models also give the most on-target first alarms (42--50\% of errorful episodes, vs.\ 23\% for SGT on the same subset), but they alarm on 30--77\% of correct episodes. Without reasoning, several models output near-zero scores at almost every step. LLMs can thus often recognize that something is off, but they still struggle to identify the step at which an error occurs and to raise a timely alarm, whereas the human reference raises an on-target alarm in 73\% of errorful episodes at a 17\% false-alarm rate.

\section{Limitations and Broader Impact}
\textsc{HelpMeOut} is symbolic: episodes are discrete action sequences over a PDDL-style state, which gives large-scale data with deterministic onsets but assumes that actions have already been recognized from video or sensors. It therefore captures neither recognition errors nor the multimodal cues through which errors manifest (hesitation, gaze, fumbling), and its results are an upper-bound proxy for symbolic-level monitoring; grounding the episodes in video or real ADL recordings is left to future work.

Two further limits follow from how episodes are generated. First, every errorful trajectory is physically executable, so failures in which an earlier lapse makes a later action impossible, such as pouring from a kettle that was never filled, are not represented. Second, correct executions are produced by a planner that completes one subgoal before starting the next; they vary in order and intermediate placements but do not interleave subgoals or contain the harmless redundancy of everyday human action, and insertion-type errors are therefore partly identifiable from surface patterns. The errors are generated rather than recorded: a label-blinded review by two clinical experts judged most generated errors plausible (supplementary material), but this is not patient-level validation.

Deployment raises risks the benchmark does not measure. False alarms interrupt the person and erode their autonomy and trust, whereas missed alarms can let an unsafe lapse, such as a stove left on, go unnoticed; thresholds should therefore be set with users and caregivers rather than by a benchmark score. In-home monitoring also requires informed consent, including from people whose capacity to consent may change, data minimization, and the person's control over what is recorded. \textsc{HelpMeOut} contains only synthetic episodes and no personal data, and results on it are not evidence that a system is safe to deploy.

\section{Conclusions}
We presented \textsc{HelpMeOut}, a benchmark that reframes household assistance as online error monitoring rather than goal inference. By pairing correct executions with neuropsychologically grounded errorful ones and asking whether, when, and what kind of error occurs, it exposes a gap that goal-inference and task-completion benchmarks miss: learned monitors, symbolic references, and frontier LLMs can often tell that an episode went wrong, especially once it has ended, but rarely detect and localize errors as they happen, and errors that require knowing which object or tool the task calls for remain near chance. We hope \textsc{HelpMeOut} serves as a testbed for assistive agents that catch execution breakdowns before a task fails. Results of further expert validation will be released on the project website. 

\clearpage


\bibliography{example}

\clearpage
\renewcommand{\topfraction}{0.92}
\renewcommand{\bottomfraction}{0.85}
\renewcommand{\textfraction}{0.07}
\renewcommand{\floatpagefraction}{0.75}
\setcounter{topnumber}{3}
\setcounter{totalnumber}{4}
\appendix
\noindent This appendix provides the additional details for the main paper.
Section~\ref{app:design} describes how the benchmark is built, from its neuropsychological grounding to the symbolic domain and the Goal-Control error
generator. Section~\ref{app:dataset} reports dataset statistics and the data split. Section~\ref{app:experiments} covers the baseline and human-study implementations together with the complete LLM results.


\section{Benchmark Design}
\label{app:design}

\subsection{Neuropsychology of Everyday Action}
\label{app:neuro}

\hmo builds its error labels and generation model on prior work in clinical neuropsychology on naturalistic action. We draw on three lines of work.

\paragraph{Performance-based instruments.}
The Naturalistic Action Test (NAT) and the Multi-Level Action Test (MLAT) ask participants to perform
familiar multi-object tasks such as making toast and coffee, wrapping a gift, packing a lunchbox---and score the resulting errors against a standardized taxonomy of omissions, sequence errors, substitutions, additions, and perseverations
\citep{schwartz2002naturalistic,schwartz1998naturalistic}.
These instruments establish that everyday-action failures are structured
and recurrent across clinical populations (left-CVA, right-CVA,
traumatic brain injury, dementia); this structure makes everyday-action errors suitable for systematic monitoring.

\paragraph{The Supervisory Attentional System (SAS) and contention scheduling.}
Contention-scheduling accounts describe routine action as competition among
activated schemas, while the Supervisory Attentional System (SAS) provides
top-down control when routine schema selection is insufficient
\citep{norman1986attention,cooper2000contention,cooper2006hierarchical}.
Within this view, weak or decaying task goals can manifest as omissions, whereas
poor suppression of competing schemas can manifest as commission errors,
especially around subgoal transitions
\citep{schwartz2006cognitive,schwartz1991quantitative}.

\begin{figure}[t]
\centering
\includegraphics[width=0.92\linewidth]{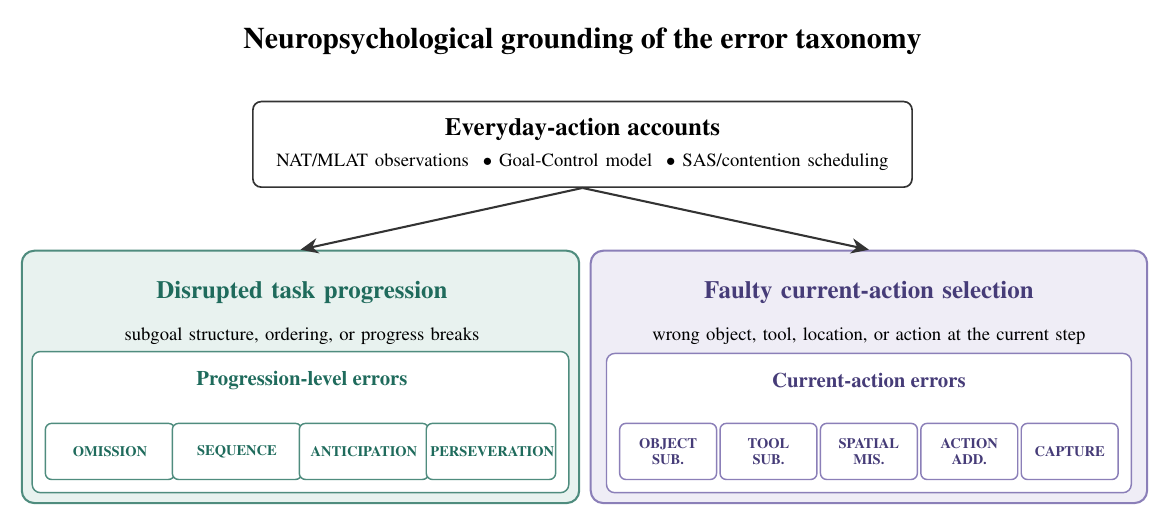}
\caption{The taxonomy is organized by what an error
disrupts: progression-level errors break the subgoal structure and ordering of
the task, whereas current-action errors corrupt the selection of the current
object, tool, location, or action. This organizing axis is distinct from the
Goal-Control generator's factorization (goal activation vs.\ executive control;
Section~\ref{app:generation}), which we use to sample errors.}
\label{fig:neuro-map}
\end{figure}

\paragraph{The Goal-Control account.}
\citet{giovannetti2021goal} synthesize these observations into a Goal-Control
model that separates failures of goal activation from failures of
executive control. We use this two-factor account in the generator
(Section~\ref{app:generation}), but the error taxonomy itself is organized by
what an error disrupts: the progression of the task or the selection of the
current action. Figure~\ref{fig:neuro-map} summarizes this organizing axis,
and the per-category clinical error rates we calibrate against are reported in
Table~\ref{tab:clinical-rates}.

\begin{table}[t]
\centering
\caption{Per-category error proportions used by the \hmo generator in the
nine-category taxonomy: (a) the omission--commission split and (b) the
commission-category weights $w_c$. We set these proportions from
everyday-action error rates reported for clinical populations
\citep{schwartz2002naturalistic,giovannetti2021goal,giovannetti2002naturalistic},
pooling the left-CVA ($n{=}16$), right-CVA ($n{=}30$), TBI ($n{=}30$),
and dementia ($n{=}54$) groups. We then make small simulator-specific
adjustments and add categories supported by the broader action-error
literature \citep{norman1986attention}.  Because an error type can be realized only when it is feasible (Section~\ref{app:generation}), the released episodes were selected per category so that the released distribution approximates these targets (Figure~\ref{fig:calibration}).}
\label{tab:clinical-rates}
\begin{tabular}{l c}
\toprule
\multicolumn{2}{l}{\emph{(a) Omission vs.\ commission (share of all errors)}} \\
\midrule
Omission                 & 0.44 \\
Commission               & 0.56 \\
\midrule
\multicolumn{2}{l}{\emph{(b) Commission category weight $w_c$ (sums to 1.00)}} \\
\midrule
Sequence                 & 0.22 \\
Object substitution      & 0.17 \\
Perseveration            & 0.15 \\
Capture                  & 0.13 \\
Spatial misreach         & 0.13 \\
Anticipation             & 0.10 \\
Tool substitution        & 0.05 \\
Action addition          & 0.05 \\
\bottomrule
\end{tabular}
\end{table}

\subsection{Object Hierarchy and Symbolic Domain}
\label{app:domain}

\hmo represents each household scene with an object-centric symbolic state. All
entities are organized into a five-class hierarchy (Table~\ref{tab:hierarchy}).
The \textsc{location} class contains the 31 non-movable entities an agent can
move to, grouped by whether objects can be placed \texttt{ontop} of them or
\texttt{inside} them; the remaining four classes contain the 156 movable object
types referred to in the main text. Including locations, the domain defines 187
object types in total; the 156 ``object types'' quoted in the main text counts
only movable objects (\textsc{receptacle}, \textsc{food}, \textsc{tool},
\textsc{thing}).

\begin{table}[t]
\centering
\caption{Object hierarchy. Five classes spanning 187 types; the 156 movable
types (\textsc{receptacle}/\textsc{food}/\textsc{tool}/\textsc{thing}) are the
``object types'' of the main text, and the 31 \textsc{location} types are the
non-movable \texttt{ontop}/\texttt{inside} locations.}
\label{tab:hierarchy}
\setlength{\tabcolsep}{5pt}
\begin{tabular}{p{2.4cm} c p{8.3cm}}
\toprule
Class & \#Types & Subclasses \\
\midrule
\textsc{location}   & 31 & \texttt{ontop}, \texttt{inside} (floor, countertop,
table, desk, shelf, stove, bed, sofa, toilet, cabinet, refrigerator, microwave,
oven, dishwasher, sink, washing\_machine, dryer, toaster, closet, drawer,
medicine\_cabinet, bathtub, \ldots) \\
\addlinespace
\textsc{receptacle} & 31 & tableware, vessel, cooking\_vessel, bag, box,
basket, tray, bucket, ashcan \\
\addlinespace
\textsc{food}       & 49 & fruit, vegetable, drink, protein, bread\_grain,
spread, condiment, sauce, prepared\_food, snack \\
\addlinespace
\textsc{tool}       & 31 & cutlery, cleaning\_tool, cleansing, wrapping\_tool,
writing\_tool, toiletry, laundry\_tool \\
\addlinespace
\textsc{thing}      & 45 & clothing, linen, soft\_furnishing, footwear,
headwear, accessory, wrapping\_material, paper\_product, personal\_item,
medication \\
\bottomrule
\end{tabular}
\end{table}

\paragraph{Predicates.}
The state is described with spatial relations (\texttt{inside},
\texttt{ontop}, \texttt{human-at}, \texttt{holding}, \texttt{hand-empty}),
boolean state attributes (\texttt{open}, \texttt{toggled}, \texttt{cooked},
\texttt{frozen}, \texttt{sliced}, \texttt{soaked}, \texttt{dusty},
\texttt{stained}, \texttt{toasted}, \texttt{buttered}, \texttt{stirred},
\texttt{folded}, \texttt{ironed}, \texttt{hung}, \texttt{worn}, \texttt{wrapped},
\texttt{packed}, \texttt{taken}, \ldots), affordance meta-predicates
(\texttt{movable}, \texttt{openable}, \texttt{toggleable}, \texttt{cookable},
\texttt{freezable}, \texttt{sliceable}, \texttt{soakable}, \texttt{foldable},
\texttt{hangable}, \texttt{wearable}, \texttt{toastable}, \ldots), and
per-type predicates (\texttt{type-knife}, \texttt{type-spoon},
\texttt{type-iron}, \ldots) used to express tool/object identity constraints.
Spatial relations are represented directly (only the immediate container/%
surface), so moving a container automatically carries its contents.

\subsection{Action Operators}
\label{app:actions}

The domain defines 38 action schemas
(Table~\ref{tab:actions}), each specified in PDDL with full preconditions and
effects; we give intuitive semantics here and release the complete domain file.

\begin{longtable}{p{4.9cm} p{8.2cm}}
\caption{The 38 action operators in \hmo. The semantics column gives an
informal description of each operator's effect.}
\label{tab:actions}\\
\toprule
Operator (\& key arguments) & Semantics \\
\midrule
\endfirsthead
\multicolumn{2}{c}{\tablename~\thetable{} (continued)}\\
\toprule
Operator (\& key arguments) & Semantics \\
\midrule
\endhead
\bottomrule
\endfoot
\texttt{human-move} \texttt{(from,to)} & Move between two non-movable locations. \\
\texttt{human-pick-up-at-location} \texttt{(obj,loc)} & Pick an object off a location (hand must be empty). \\
\texttt{human-pick-up-from-\allowbreak receptacle-at-location} \texttt{(obj,rec,loc)} & Pick an object out of/off a receptacle at a location. \\
\texttt{human-put-inside-location} \texttt{(obj,loc)} & Put the held object inside a location. \\
\texttt{human-put-ontop-location} \texttt{(obj,loc)} & Put the held object on top of a location. \\
\texttt{human-put-inside-\allowbreak receptacle-at-location} \texttt{(obj,rec,loc)} & Put the held object inside a receptacle. \\
\texttt{human-put-ontop-\allowbreak receptacle-at-location} \texttt{(obj,rec,loc)} & Put the held object on top of a receptacle. \\
\texttt{human-open-location} \texttt{(loc)} & Open an openable location. \\
\texttt{human-close-location} \texttt{(loc)} & Close an open location. \\
\texttt{human-open-\allowbreak receptacle-at-location} \texttt{(rec,loc)} & Open an openable receptacle. \\
\texttt{human-close-\allowbreak receptacle-at-location} \texttt{(rec,loc)} & Close an open receptacle. \\
\texttt{human-toggle-on-location} \texttt{(loc)} & Toggle a toggleable appliance on. \\
\texttt{human-toggle-off-location} \texttt{(loc)} & Toggle a toggleable appliance off. \\
\texttt{human-heat-obj} \texttt{(obj,loc)} & Heat/cook an object at a microwave/oven. \\
\texttt{human-heat-obj-on-stove} \texttt{(obj,vessel,loc)} & Cook an object in a vessel on a stove. \\
\texttt{human-cool-obj} \texttt{(obj,loc)} & Freeze an object inside a refrigerator. \\
\texttt{human-slice-obj} \texttt{(obj,tool,loc)} & Slice a sliceable object with a knife. \\
\texttt{human-soak-obj} \texttt{(obj,loc)} & Soak a soakable object at a sink. \\
\texttt{human-clean-obj-at-location} \texttt{(obj,tool,loc)} & Remove dust/stain from an object with a cleaning tool. \\
\texttt{human-clean-location} \texttt{(tool,loc)} & Remove dust/stain from a location. \\
\texttt{human-discard-obj} \texttt{(obj,bin,loc)} & Throw an object into a bin. \\
\texttt{human-replace-obj} \texttt{(new,old,loc)} & Swap a new object in for an old one on a surface. \\
\texttt{human-toast-obj} \texttt{(obj,loc)} & Toast a toastable object (sets toasted, cooked). \\
\texttt{human-spread} \texttt{(obj,substance,tool,loc)} & Spread a substance (butter/jam) onto an object with a tool. \\
\texttt{human-pour} \texttt{(content,source,dest,loc)} & Pour content from a source vessel into a destination. \\
\texttt{human-stir} \texttt{(container,tool,loc)} & Stir a container with a spoon. \\
\texttt{human-wrap-obj} \texttt{(obj,material,tool,loc)} & Wrap an object in wrapping material. \\
\texttt{human-close-gift-box-flaps} \texttt{(box,loc)} & Close the flaps of a gift box. \\
\texttt{human-close-gift-box-lid} \texttt{(box,loc)} & Close the lid of a gift box. \\
\texttt{human-pack-obj} \texttt{(obj,container,loc)} & Pack an object into a container. \\
\texttt{human-wear-obj} \texttt{(obj)} & Put on a wearable object. \\
\texttt{human-remove-worn-obj} \texttt{(obj)} & Take off a worn object. \\
\texttt{human-fold-obj} \texttt{(obj,loc)} & Fold a foldable object. \\
\texttt{human-iron-obj} \texttt{(obj,tool,loc)} & Iron an object with an iron. \\
\texttt{human-hang-obj} \texttt{(obj,loc)} & Hang a hangable object. \\
\texttt{human-wash-clothes} \texttt{(obj,loc)} & Wash clothing at a washing machine. \\
\texttt{human-dry-clothes} \texttt{(obj,loc)} & Dry clothing at a dryer. \\
\texttt{human-take-medication} \texttt{(obj)} & Take a medication object. \\
\end{longtable}

\paragraph{Action constraints.}
Several operators carry tool/location constraints enforced through the
per-type predicates: slicing requires a \texttt{type-knife} tool, stirring a
\texttt{type-spoon} tool, ironing a \texttt{type-iron} tool; heating is only
valid at a microwave/oven/stove, cooling only inside a refrigerator, soaking
only at a sink, and cleaning only with a suitable cleaning tool. These
constraints are exactly the preconditions whose violation defines the
\textsc{tool\_substitution} error class (Section~\ref{app:taxonomy}).

\subsection{Task Templates}
\label{app:templates}

\begin{figure}[H]
\centering
\includegraphics[width=\linewidth]{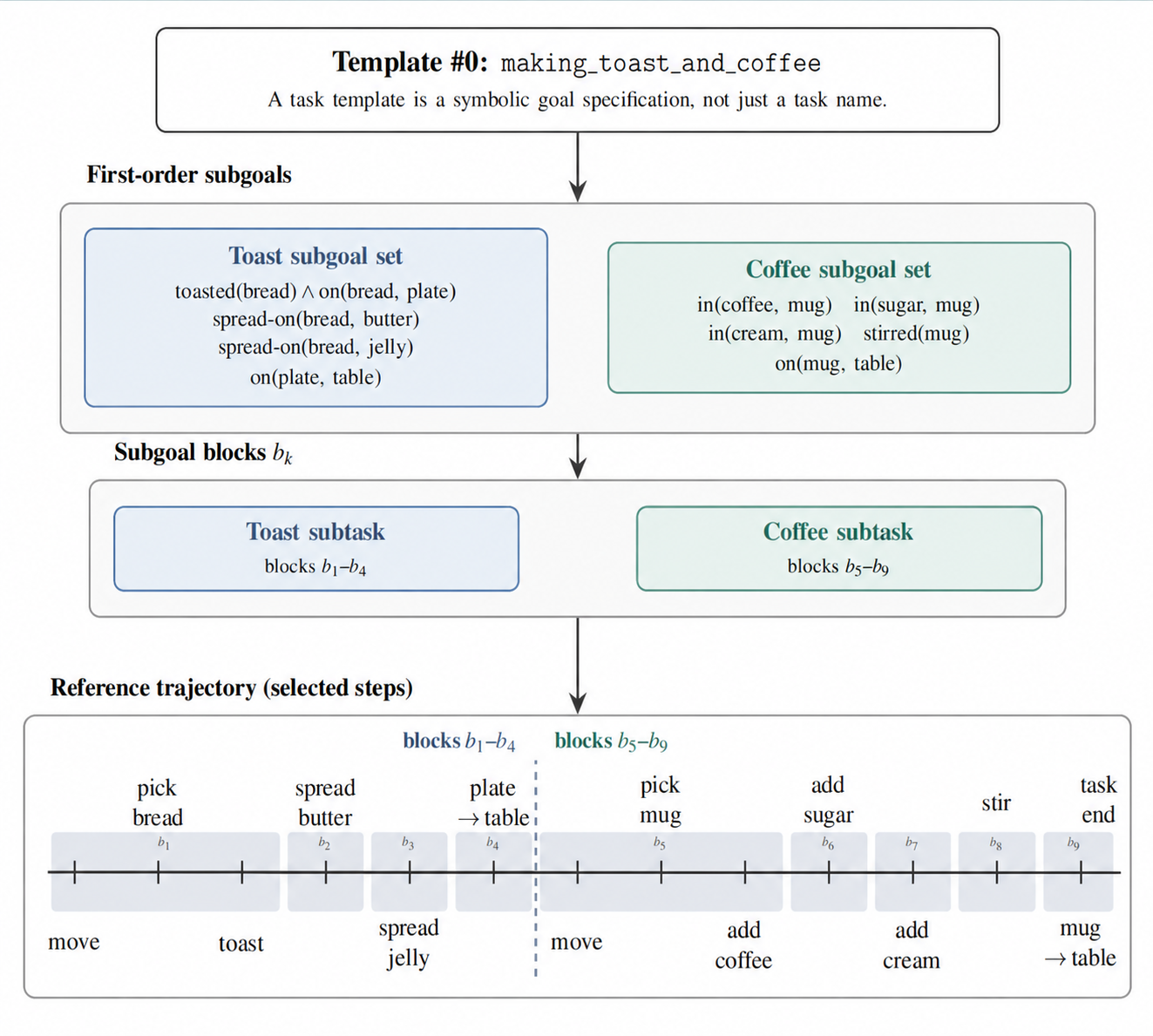}
\caption{An example of one executable task template. Template \#0
(\texttt{making\_toast\_and\_coffee}) expands into first-order subgoals---the
subgoal blocks $b_k$, here grouped into a toast and a coffee subtask---which a
planner solves into a correct reference trajectory (the shaded bands are the
$b_k$). The block boundary shown on the trajectory is the structure the
generator uses for progression-level errors; in generated episodes the block order is randomized, and onsets are labelled by the category rules of Section~\ref{app:taxonomy}. Only representative actions are shown.}
\label{fig:template-example}
\end{figure}

\hmo spans 50 ADL/IADL task templates organized into eight families
(Table~\ref{tab:templates}). Each template is anchored in a clinical
assessment, functional scale, or occupational-therapy framework for everyday
action---NAT, PASS, the Lawton IADL scale, the Katz ADL index, the Executive
Function Performance Test (EFPT), or the OTPF-4 occupational-therapy
framework---and is annotated with a complexity tier
(simple / medium / complex) reflecting the number of accomplishment steps. Each
template is expressed as a set of first-order subgoals over the predicates of
Section~\ref{app:domain}; Figure~\ref{fig:template-example} shows a worked
example.

{\small
\setlength{\tabcolsep}{4pt}
\renewcommand{\arraystretch}{1.05}
\begin{longtable}{c l l l}
\caption{The 50 ADL/IADL task templates, grouped into eight families.}
\label{tab:templates}\\
\toprule
\# & Template & Anchor & Complexity \\
\midrule
\endfirsthead
\multicolumn{4}{c}{\tablename~\thetable{} (continued)}\\
\toprule
\# & Template & Anchor & Complexity \\
\midrule
\endhead
\bottomrule
\endfoot
\multicolumn{4}{l}{\textit{F1: Food preparation}}\\
0  & making\_toast\_and\_coffee              & NAT         & complex \\
1  & slicing\_fruit\_and\_refrigerating       & NAT         & medium  \\
2  & cooking\_and\_plating\_protein           & NAT         & simple  \\
3  & storing\_fruits\_and\_vegetables & Lawton IADL & simple \\
4  & making\_a\_salad                         & NAT         & medium  \\
5  & preparing\_breakfast                     & NAT         & complex \\
\addlinespace
\multicolumn{4}{l}{\textit{F2: Gift wrapping}}\\
6  & wrapping\_a\_book                        & PASS        & simple  \\
7  & wrapping\_two\_items                     & PASS        & simple  \\
8  & wrapping\_book\_and\_boxing              & PASS        & medium  \\
9  & wrapping\_shirt\_and\_bagging            & PASS        & medium  \\
\addlinespace
\multicolumn{4}{l}{\textit{F3: Packing}}\\
10 & packing\_school\_bag\_and\_lunch         & Lawton IADL & complex \\
\addlinespace
\multicolumn{4}{l}{\textit{F4: Table setting \& serving}}\\
11 & setting\_table\_and\_serving\_meal       & PASS        & complex \\
12 & setting\_a\_simple\_table                & PASS        & simple  \\
13 & plating\_a\_cooked\_dish                 & OTPF-4      & simple  \\
14 & preparing\_a\_tea\_tray                  & Lawton IADL & medium  \\
15 & setting\_table\_for\_two                 & PASS        & medium  \\
16 & serving\_breakfast\_on\_tray             & OTPF-4      & medium  \\
17 & clearing\_the\_dining\_table             & Lawton IADL & medium  \\
\addlinespace
\multicolumn{4}{l}{\textit{F5: Cleaning \& tidying}}\\
18 & cleaning\_kitchen\_and\_storing\_dishes  & PASS        & complex \\
19 & cleaning\_bathroom                       & OTPF-4      & medium  \\
20 & cleaning\_dining\_room                   & OTPF-4      & medium  \\
21 & wiping\_a\_countertop                    & PASS        & simple  \\
22 & washing\_dishes\_and\_putting\_away      & Lawton IADL & simple  \\
23 & doing\_dishes\_and\_cleaning\_sink       & PASS        & medium  \\
24 & tidying\_a\_desk                         & OTPF-4      & medium  \\
25 & organizing\_a\_bookshelf                 & OTPF-4      & medium  \\
26 & cleaning\_living\_room                   & Lawton IADL & medium  \\
27 & tidying\_a\_bedroom                      & OTPF-4      & complex \\
\addlinespace
\multicolumn{4}{l}{\textit{F6: Laundry \& clothing care}}\\
28 & folding\_and\_storing\_clothes           & Lawton IADL & medium  \\
29 & hanging\_clothes\_and\_folding\_linens   & Lawton IADL & medium  \\
30 & ironing\_folding\_storing\_shirt         & Lawton IADL & medium  \\
31 & folding\_and\_storing\_linens            & Lawton IADL & medium  \\
32 & folding\_clothes\_and\_ironing           & Lawton IADL & medium  \\
33 & putting\_away\_folded\_clothes           & Lawton IADL & simple  \\
34 & hanging\_a\_shirt\_to\_dry               & Lawton IADL & simple  \\
35 & sorting\_laundry\_into\_baskets          & Lawton IADL & medium  \\
\addlinespace
\multicolumn{4}{l}{\textit{F7: Medication \& personal care}}\\
36 & preparing\_and\_taking\_medication       & EFPT        & medium  \\
37 & taking\_a\_pill                          & EFPT        & simple  \\
38 & filling\_a\_glass\_of\_water             & PASS        & simple  \\
39 & organizing\_weekly\_pill\_box            & EFPT        & medium  \\
40 & setting\_up\_toiletries                  & PASS        & medium  \\
41 & restocking\_bathroom\_supplies           & OTPF-4      & complex \\
\addlinespace
\multicolumn{4}{l}{\textit{F8: Daily routines}}\\
42 & preparing\_to\_leave\_home               & OTPF-4      & medium  \\
43 & putting\_on\_shoes\_and\_grabbing\_keys  & Katz ADL    & simple  \\
44 & getting\_dressed                         & Katz ADL    & simple  \\
45 & dressing\_for\_rain                      & OTPF-4      & medium  \\
46 & preparing\_for\_work                     & Lawton IADL & complex \\
47 & packing\_a\_gym\_bag                     & OTPF-4      & medium  \\
48 & coming\_home\_routine                    & OTPF-4      & medium  \\
49 & preparing\_for\_bed                      & OTPF-4      & medium  \\
\end{longtable}
}

\subsection{Error Taxonomy}
\label{app:taxonomy}

\begin{figure}[t]
\centering
\includegraphics[width=\linewidth]{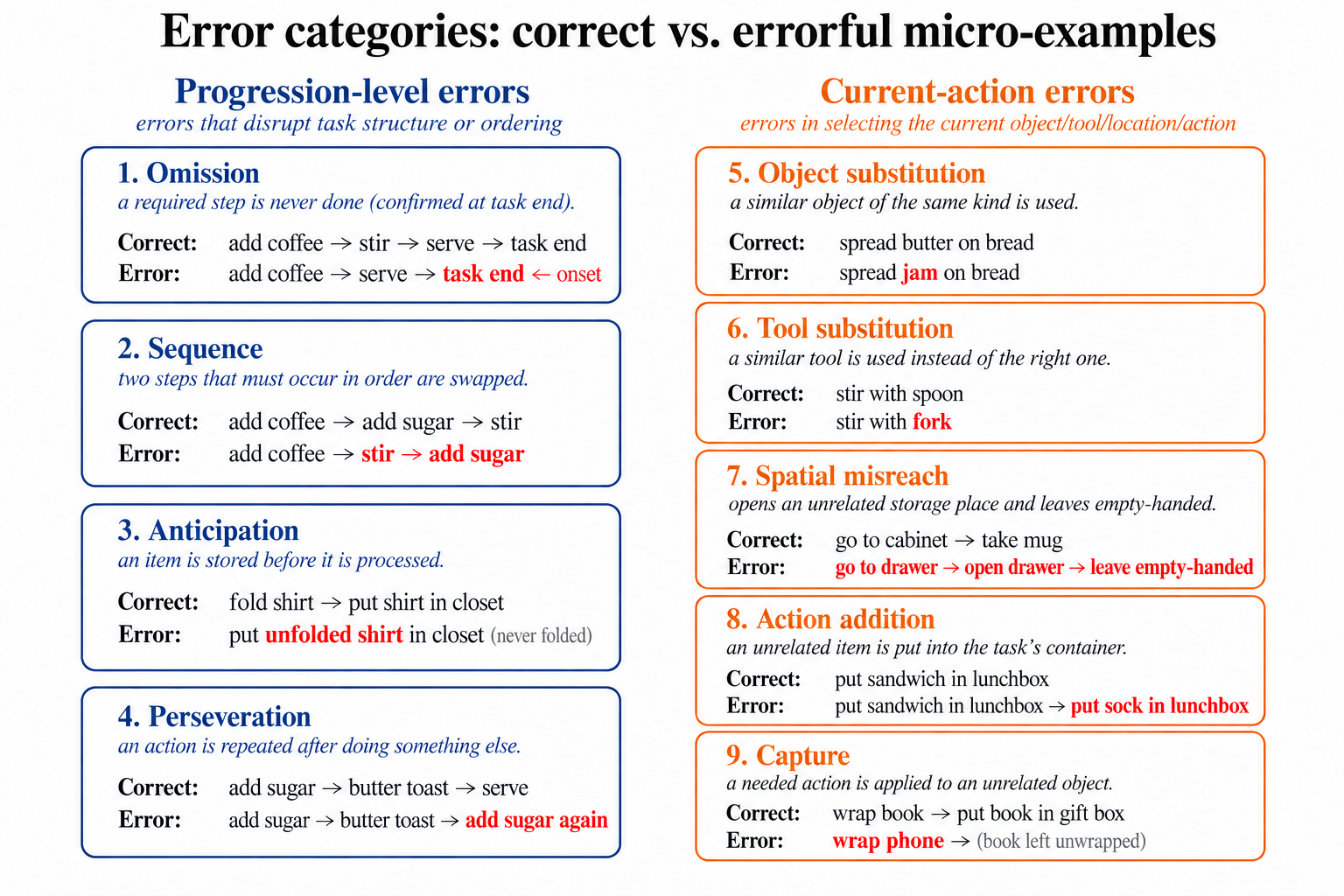}
\caption{A correct-vs-errorful micro-example for each of the nine error
categories, grouped into progression-level and current-action errors. In every
\textsc{error} row, the step(s) that diverge from the correct execution are
highlighted in red. Examples use the household domain of
Figure~\ref{fig:template-example}.}
\label{fig:error-examples}
\end{figure}

The nine error categories are organized into two levels. Progression-%
level errors disrupt the structure of task execution; current-action
errors affect local action selection, object use, or spatial targeting. For
each category we give its definition, trigger condition, and where the onset is
labelled; Figure~\ref{fig:error-examples} pairs a correct and an errorful
micro-example per category.

Each definition below names the step at which the onset is labelled; onsets are recognizable from the observed prefix and the task, without the reference execution. An immediately repeated action without additional effect, or walking back and forth between places the task uses, is harmless redundancy and not an error.

\paragraph{Progression-level errors.}
\begin{itemize}
\item \textbf{\textsc{Omission}.} An entire subgoal block is deleted. Because correct executions may perform subgoals in different orders, the missing subgoal could still be performed later; the omission is therefore confirmed at the task-end event. \emph{Onset:} the task-end step. When the missing step has an earlier observable consequence, such as storing an item before it is processed, the error is labelled \textsc{anticipation} and the deleted step is not counted as an omission, following MLAT coding~\citep{schwartz1998naturalistic}.
\item \textbf{\textsc{Sequence}.} Two steps that must occur in a specific order are performed in the wrong order. Two constructions are used: stirring before the last ingredient is added (the MLAT example ``stir mug of water, then add grinds''), and violating a step-order norm (buttering bread before toasting it, folding before ironing, taking a pill before pouring water, putting on a jacket before the required shirt). Other orders that reach the goal are correct. \emph{Onset:} the step performed out of order.
\item \textbf{\textsc{Anticipation}.} An item is stored, served, or packed before it receives the processing the task requires, or a gift box's lid is closed without first closing its flaps, which are never closed afterwards. \emph{Onset:} the step that commits the unprocessed item: putting it into its final place, closing the container that holds it, or delivering that container to its final place.
\item \textbf{\textsc{Perseveration}.} After doing something else in between, the person repeats an action that should be done only once, such as adding sugar to the coffee again after buttering the toast, or taking an extra dose of a pill the task does not require. \emph{Onset:} the repeated action.
\end{itemize}

\paragraph{Current-action errors.}
\begin{itemize}
\item \textbf{\textsc{Object\_substitution}.} A confusable object of the same subclass is used in place of the object the task needs (e.g.\ jam for butter). \emph{Onset:} the first step that uses the substitute.
\item \textbf{\textsc{Tool\_substitution}.} A confusable tool is used in place of the appropriate one (e.g.\ a fork where a spoon is required). \emph{Onset:} the first step that uses the substitute tool.
\item \textbf{\textsc{Action\_addition}.} An item unrelated to the task (not named in the goal and of a different subclass from every task object) is put into or onto the task's target container, plate, or tray. Removing it later does not undo the error. \emph{Onset:} the step that puts it there.
\item \textbf{\textsc{Capture}.} An action the task needs is applied to an unrelated object of a different subclass instead of the task's object, which is then left unprocessed. \emph{Onset:} the action applied to the unrelated object (fetching it or walking to it does not count).
\item \textbf{\textsc{Spatial\_misreach}.} The person goes to a storage location that the task never uses, opens it, and leaves without taking anything. \emph{Onset:} the step of leaving that location empty-handed, the first step at which the visit is recognizably unrelated to the task.
\end{itemize}

\subsection{Error-Generation Model}
\label{app:generation}

\paragraph{Goal-Control process (supplementary parameter values).}
The main text defines the Goal-Control process (Eq.~2): for each subgoal block
$b_k$ an activation
$u_k=\theta_{\mathrm{goal}}(1-\theta_{\mathrm{decay}})^k+\epsilon_k$ gates
omission when $u_k<\tau_{\mathrm{omit}}$, and surviving blocks receive a
commission with probability $\theta_{\mathrm{ctrl}}$. Here we report only the
values deferred to this supplement: the omission threshold
$\tau_{\mathrm{omit}}{=}0.3$ and the activation noise $\sigma{=}0.1$
(with $\epsilon_k\sim\mathcal{N}(0,\sigma^2)$). The impaired profile
$(\theta_{\mathrm{goal}},\theta_{\mathrm{decay}},\theta_{\mathrm{ctrl}})
{=}(0.48,0.20,0.12)$ and the omission-vs-commission split it targets are
reported in Table~\ref{tab:clinical-rates}; all generator hyperparameters are
collected in Table~\ref{tab:hparams}.

\begin{table}[t]
\centering
\caption{Generator hyperparameters.}
\label{tab:hparams}
\begin{tabular}{lcl}
\toprule
Symbol / name & Value & Meaning \\
\midrule
$\theta_{\mathrm{goal}}$            & 0.48 & baseline goal salience \\
$\theta_{\mathrm{decay}}$           & 0.20 & per-subgoal activation decay (cognitive load) \\
$\theta_{\mathrm{ctrl}}$            & 0.12 & commission-error probability \\
$\tau_{\mathrm{omit}}$ (\texttt{OMIT\_THRESHOLD})       & 0.30 & omission activation threshold \\
$\sigma$ (\texttt{ACTIVATION\_NOISE\_STD})              & 0.10 & activation noise std.\ dev. \\
min.\ trajectory length             & 10   & shorter trajectories are filtered \\
\bottomrule
\end{tabular}
\end{table}

\paragraph{Two-axis per-subgoal decision.}
The generator walks the subgoal blocks of the reference trajectory in order and
applies two decisions at each block. \emph{(i) Goal axis:} it first tests the
activation $u_k$; if $u_k<\tau_{\mathrm{omit}}$ the block is dropped as an
\textsc{omission}. \emph{(ii) Control axis:} otherwise, with probability
$\theta_{\mathrm{ctrl}}$ a commission is injected, and its category is drawn from
the weights $w_c$ (Table~\ref{tab:clinical-rates}(b)) restricted to the
categories that are feasible at the current step. These two axes correspond to
the two panels of Table~\ref{tab:clinical-rates}: the goal-vs-control decision
gives the omission-vs-commission split~(a), and the weighted draw on the control
axis gives the commission-category weights~(b). Because a commission category is
realized only when feasible in the current state, the dataset frequencies can
differ from $w_c$ (see \emph{Per-category feasibility} below).

\paragraph{Per-category feasibility $\phi_c(s_k)$.}
Not every commission category can be injected at every step. Each injector
requires a particular trajectory or scene configuration: for example,
\textsc{object\_substitution} requires a confusable alternative object in the
scene, while \textsc{sequence} requires reorderable subgoal blocks. We represent
this with a binary feasibility gate $\phi_c(s_k)$, which is 1 only when category
$c$ can be realized at step $k$. As a result, the observed category frequencies
are shaped by both the sampling weights $w_c$ and the states in which each error
is feasible. Table~\ref{tab:feasibility} summarizes these gates. 
 Every injected episode must also pass the validation of Section~\ref{app:validation}, match its category rule exactly once at the injected step, and the reference execution must not match the rule.

\begin{table}[t]
\centering
\caption{Per-category feasibility gate $\phi_c(s_k)$: the structural condition
under which each error type can be injected at step $k$. A sampled error is
realized only when its condition holds in the current state; otherwise the
generator resamples.}
\label{tab:feasibility}
\setlength{\tabcolsep}{6pt}
\begin{tabular}{l p{9.6cm}}
\toprule
Category & Feasible when \ldots \\
\midrule
\textsc{omission}             & a subgoal block can be deleted while the trajectory stays executable \\
\textsc{sequence}             & a stir follows several additions to one container, or a step-order norm applies \\
\textsc{anticipation}         & the task requires processing an item before it is stored, served, or packed, or requires closing a gift box's flaps before its lid \\
\textsc{perseveration}        & an addition (or dose) can be repeated after intervening actions, with all goals still met \\
\textsc{object\_substitution} & a same-subclass confusable object is present in the scene \\
\textsc{tool\_substitution}   & a confusable alternative tool is present in the scene \\
\textsc{spatial\_misreach}    & an openable storage location exists that the task never uses \\
\textsc{action\_addition}     & an item unrelated to the task can be put into the task's target container \\
\textsc{capture\_error}       & an unrelated object of a different subclass affords a processing action the task needs \\
\bottomrule
\end{tabular}
\end{table}

\begin{figure}[t]
\centering
\includegraphics[width=\linewidth]{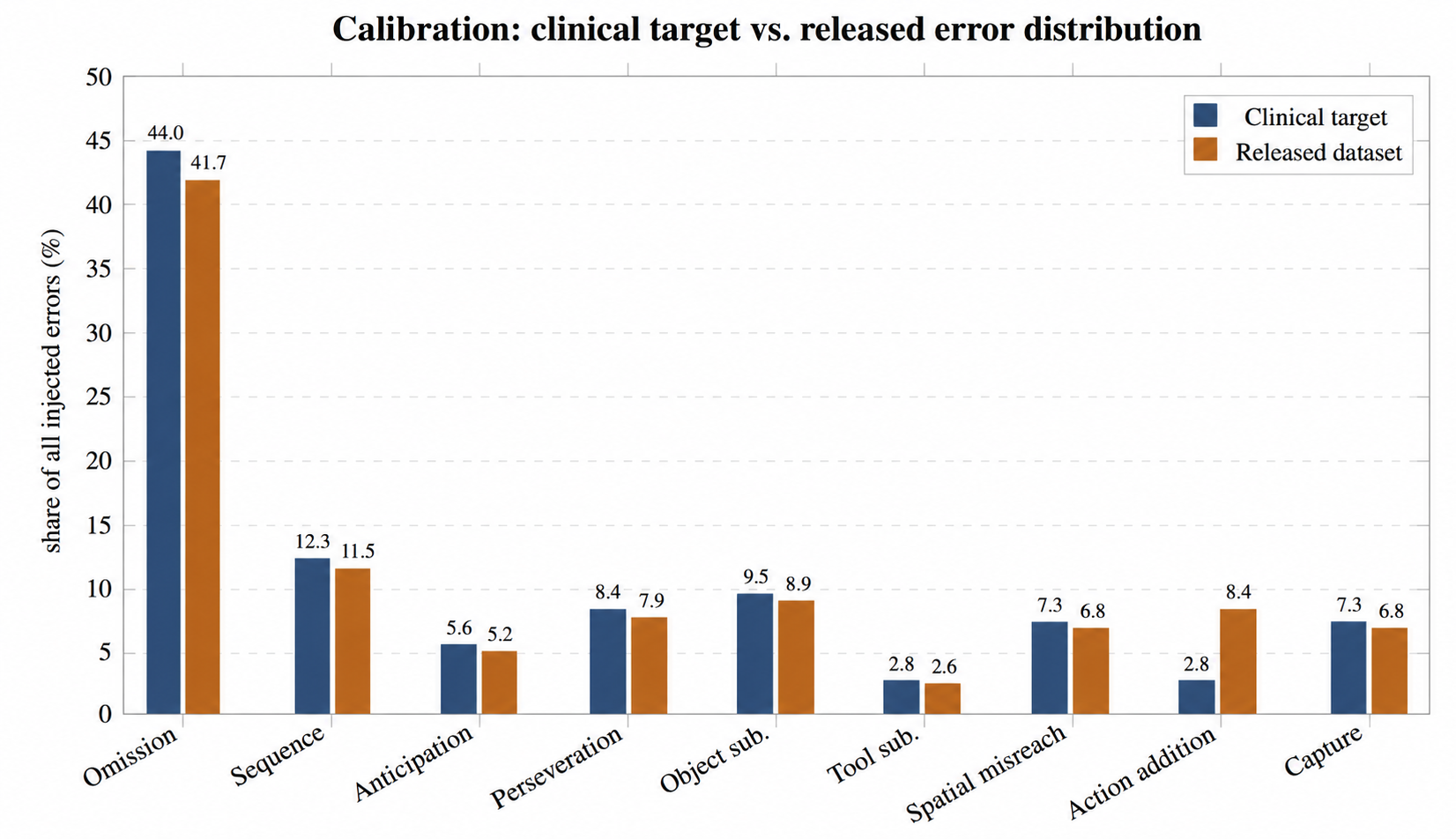}
\caption{Calibration check: the generator targets the clinical per-category
rates of Table~\ref{tab:clinical-rates} (blue); the released distribution (orange) is within 2.3 percentage points of the target for every category except action addition (8.4\% vs.\ 2.8\%), which frequently occurs as the second error of a multi-error episode. Shares are over all injected errors (12{,}817 in the release) and each series sums to 100\%.}
\label{fig:calibration}
\end{figure}

\paragraph{Calibration.}
We calibrate $\theta_{\mathrm{goal}},\theta_{\mathrm{decay}},
\theta_{\mathrm{ctrl}}$ and $w_c$ so that the marginal error distribution
produced by the generator matches the per-category rates of
Table~\ref{tab:clinical-rates}. Figure~\ref{fig:calibration} compares the
generated distribution against the clinical targets. Because the feasibility gates bind unevenly, the raw generator output under-represents low-feasibility categories; the released episodes were therefore selected per category from a larger generated pool, so that the released distribution approximates the targets.

\subsection{Dual-Standard Validation}
\label{app:validation}

A core requirement is that errorful trajectories be \emph{semantically wrong yet
physically executable}. We therefore validate with two standards.

\paragraph{Strict standard (7 categories).}
\textsc{omission}, \textsc{sequence}, \textsc{anticipation},
\textsc{perseveration}, \textsc{action\_addition}, \textsc{capture\_error}, and
\textsc{spatial\_misreach} must pass a complete PDDL replay: every action's
preconditions must hold in the simulated state. These errors are
``legal-but-wrong'' executions and remain fully physically consistent.

\paragraph{Relaxed standard (2 categories).}
\textsc{tool\_substitution} and \textsc{object\_substitution} would be rejected
by a naive validator---using the wrong tool/object is the error. For
these we enforce only the physical chain---navigation, accessibility,
containment, and hand occupancy---while relaxing the single semantic
precondition that defines the substitution (e.g.\ the \texttt{type-spoon}
requirement of \texttt{human-stir}). Within the error block, only the semantic
type constraint is relaxed; all physical preconditions still hold.

\paragraph{Confusable-alternate selection.}
Substituted tools/objects are drawn from a hierarchy-constrained confusable set:
\textsc{tool}/\textsc{food}/\textsc{thing} substitutions stay within the same
subclass, \textsc{receptacle} substitutions may cross subclass, and no
substitution ever crosses a parent class. This keeps each substitution
interpretable as a human confusion rather than an arbitrary swap.

\section{Dataset Statistics}
\label{app:dataset}

\subsection{Corpus Overview}
\label{app:stats}
\begin{figure}[t]
\centering
\includegraphics[width=\linewidth]{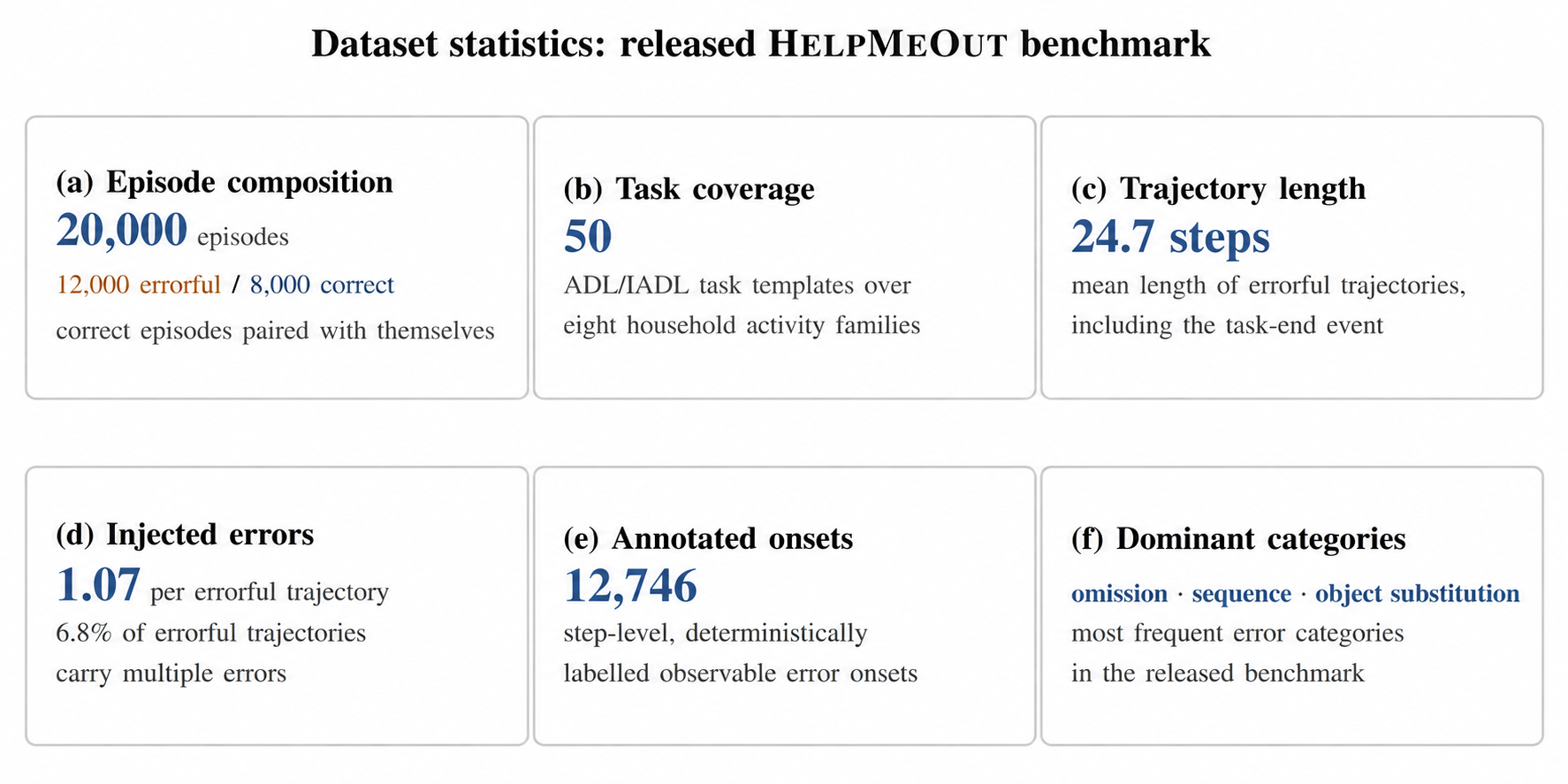}
\caption{Summary statistics of the released \hmo benchmark: (a) episode
composition (paired errorful/correct), (b) task coverage, (c) mean trajectory
length, (d) injected errors per trajectory, (e) annotated step-level onsets, and
(f) the most frequent error categories.}
\label{fig:stats}
\end{figure}

Figure~\ref{fig:stats} summarizes the released \hmo corpus---its episode
composition, task coverage, trajectory length, injected-error rate, annotated
onsets, and dominant error categories.

\subsection{Data Split}
\label{app:split}

The released corpus contains 20{,}000 episodes over 50 templates: 17{,}478 for training, 2{,}022 for validation, and 500 for testing. The errorful test episodes are stratified by category in proportion to the full corpus and the test set covers all 50 templates (312 errorful, 188 correct; 121 simple, 291 medium, and 88 complex episodes by template tier). All learned and non-learned baselines are evaluated on the full test set. Because frontier-LLM evaluation is substantially more expensive, LLMs and the human reference are evaluated on a fixed 60-episode subset of the test set with 30 errorful episodes, covering every category (three to four per category), and 30 correct episodes from 30 different templates; the learned and non-learned baselines are re-scored on the same subset for comparison.

\begin{table}[t]
\centering
\caption{Evaluation splits used in the main paper. The 60-episode LLM subset is
drawn from the held-out test set; all other baselines are reported both on the
full test set and on this common subset for fair comparison with LLMs.}
\label{tab:data-split}
\begin{tabular}{l c p{7.2cm}}
\toprule
Split & Episodes & Use \\
\midrule
Released corpus & 20{,}000 & Full benchmark release \\
Training / validation & 17{,}478 / 2{,}022 & Training and checkpoint selection \\
Held-out test set & 500 & Main baseline evaluation \\
LLM subset & 60 & Frontier-LLM and human-reference evaluation and matched baseline subset scores \\
\bottomrule
\end{tabular}
\end{table}

\section{Experimental Details}
\label{app:experiments}

\subsection{Learned Baselines}
\label{app:supervised}
The learned baselines are trained as discriminative error monitors on the benchmark's training corpus (disjoint from the held-out test set) and run as online monitors on the 500-episode test set. Every learned baseline receives the same input as the LLMs: the task name, the initial scene, and the action prefix; none receives the formal goal, the reference execution, or any label.

\paragraph{Seq2Seq.}
At each step, the text-only baseline encodes the task description, the initial
scene description, and the observed action prefix as token sequences. The shared
sequence representation feeds three supervised prediction heads: a step-level
error head producing $p_t$ (the reported \texttt{p\_instant} score), a 9-class
multi-label error-type head producing \texttt{p\_type}, and an auxiliary
prefix-error head used only as training supervision. The auxiliary prefix head is
not used by the reported metrics.

Training minimizes the sum of binary cross-entropy for the step-level error
head, multi-label binary cross-entropy for the type head, and auxiliary binary
cross-entropy for the prefix head.  The step-level and prefix heads are trained without positive-class weights; the type head uses per-class positive weights equal to the negative-to-positive ratio. We train with batch size 32 (two accumulated batches of 16), learning rate $10^{-4}$, and a single seed (42), evaluate on the validation split every 2{,}000 steps, and keep the checkpoint with the highest validation M2 AUPRC; the test set is scored once. 

The reported metrics are derived directly from these outputs: M1 uses
$\max_t p_t$, M2 is the AUPRC of $p_t$ against the step label $y_t$, M3 is the
first time step with $p_t \geq \theta$, and M4 uses the 9-class
\texttt{p\_type} predictions.

\paragraph{Seq2Seq+SG.}
The +SG variant augments the same Seq2Seq backbone with an explicit scene-graph
encoder. A relational graph is built from the symbolic scene state after replaying the observed prefix in the PDDL simulator (object identity, location, state, and containment; there is no geometry, hence the renaming from ``+3D'')
(\texttt{build\_local\_graph}); each node carries a 134-dimensional feature
(a 32-d learned type embedding, plus class, subclass, affordance, and
state-boolean one-hot blocks), and a two-layer R-GCN
(\texttt{RGCNConv(134,64)}\,$\to$\,ReLU\,$\to$\,Dropout(0.2)\,$\to$\,%
\texttt{RGCNConv(64,64)}\,$\to$\,ReLU) with four relation types and mean-pool
readout produces a graph-level embedding that is fused before the prediction heads. Training otherwise matches the Seq2Seq baseline (same discriminative multi-head loss). This isolates the contribution of explicit object/%
spatial state to error monitoring. Both supervised baselines serve as reference
implementations rather than new model contributions. Seq2Seq+SG performs slightly better than Seq2Seq (main paper, Table~2).

\paragraph{Per-class error-type diagnosis (M4).}

Table~\ref{tab:sup-m4} reports per-class M4 F1 of the learned monitors on the 500-episode test set. M4 scores the cumulative category predictions in errorful episodes from the first onset onward and in correct episodes at every evaluated step, where naming any category is a false positive; each episode is weighted equally. The type heads of the learned monitors name an error category in most correct episodes (at the last step: all 188 for Seq2Seq and Seq2Seq+SG, 160 for SGT), mostly omission and object substitution, which lowers their macro-F1 from 0.63--0.68 when scored on errorful episodes alone to 0.50--0.55. Omission and sequence remain the best-diagnosed categories (F1 $\geq$ 0.74). SGT is best on anticipation, perseveration, and spatial misreach, the Seq2Seq models on capture and action addition; object and tool substitution stay at or below 0.20 for all three models. Predicting categories from the task name alone, using their training frequencies per template, reaches at most 0.20 in the errorful-only scoring, so the type heads use the observed actions rather than a task prior.

\begin{table}[t]
\centering
\caption{Per-class M4 F1 ($\theta{=}0.5$) of the learned monitors on the 500-episode test set, with correct episodes included (true category set empty). The last row gives macro-F1 when only errorful episodes are scored.}
\label{tab:sup-m4}
\setlength{\tabcolsep}{6pt}
\begin{tabular}{lccc}
\toprule
Error category & Seq2Seq & Seq2Seq+SG & SGT \\
\midrule
Omission               & 0.893 & 0.888 & 0.916 \\
Sequence               & 0.778 & 0.741 & 0.921 \\
Anticipation           & 0.386 & 0.422 & 0.607 \\
Perseveration          & 0.632 & 0.664 & 0.803 \\
Object substitution    & 0.145 & 0.166 & 0.196 \\
Tool substitution      & 0.113 & 0.122 & 0.000 \\
Action addition        & 0.411 & 0.691 & 0.216 \\
Capture                & 0.708 & 0.982 & 0.502 \\
Spatial misreach       & 0.456 & 0.254 & 0.715 \\
\midrule
\textbf{macro-F1} & 0.502 & 0.548 & 0.542 \\
macro-F1, errorful episodes only & 0.631 & 0.678 & 0.638 \\
\bottomrule
\end{tabular}
\end{table}

\paragraph{Per-class step-level detection (M2).}
Table~\ref{tab:m2_per_class} breaks the step-level AUPRC of the learned monitors down by error category. For each category we pool the evaluation-window steps of the test episodes of that category with those of all 188 correct episodes; for omissions all evaluated steps are scored, with the task-end step as the positive, and for all other categories the steps before task end. SGT's advantage concentrates on categories with a local action signature---\textsc{sequence} (0.814), \textsc{spatial\_misreach} (0.771), and \textsc{perseveration} (0.502)---which it can recognize from the current action and the scene graph that precedes it. Categories whose detection requires knowing which object, tool, or item the task calls for remain near chance for all three models: \textsc{object\_substitution} (0.012--0.038), \textsc{tool\_substitution} (0.003--0.020), and \textsc{action\_addition} (0.005--0.029). The Seq2Seq models are strong only on omissions at the task-end step.

\begin{table}[t]
\centering
\small
\caption{Per-class step-level AUPRC (M2) of the learned monitors on the 500-episode test set. Each row pools the test episodes whose primary category is the given one with the 188 correct episodes; $n$ is the number of episodes and of error steps (multi-error episodes contribute all their error steps). Random equals the positive-step rate of the row.}
\label{tab:m2_per_class}
\begin{tabular}{lccccc}
\toprule
Error category & $n$ (episodes / steps) & Random & Seq2Seq & Seq2Seq+SG & SGT \\
\midrule
Omission & 137 / 137 & 0.026 & 0.696 & 0.716 & 0.730 \\
Sequence                 & 38 / 38   & 0.010 & 0.186 & 0.215 & 0.814 \\
Anticipation             & 17 / 17   & 0.005 & 0.023 & 0.085 & 0.359 \\
Perseveration            & 26 / 26   & 0.007 & 0.093 & 0.095 & 0.502 \\
Object substitution      & 30 / 30   & 0.008 & 0.018 & 0.012 & 0.038 \\
Tool substitution        & 9 / 9     & 0.003 & 0.003 & 0.003 & 0.020 \\
Action addition          & 9 / 9     & 0.003 & 0.005 & 0.024 & 0.029 \\
Capture                  & 23 / 44   & 0.012 & 0.204 & 0.189 & 0.178 \\
Spatial misreach         & 23 / 23   & 0.007 & 0.108 & 0.052 & 0.771 \\
\bottomrule
\end{tabular}
\end{table}

\subsection{Scene-Graph Transformer and CUSUM}
\label{app:sgt-cusum}
\paragraph{Scene-graph transformer (SGT).} At each step $t$, SGT encodes the task name, the symbolic scene graph obtained by replaying actions $a_{0:t-1}$ in the PDDL simulator, and the current action $a_t$; a causally masked transformer encoder (model width 128, four heads) runs over the steps, and three heads predict the step-level error probability $p_t$, an auxiliary prefix probability, and the nine type probabilities. Because the graph precedes the action and the action is a separate input, each step can be scored by whether the current action fits the current state. Training uses losses analogous to Seq2Seq (the type loss is applied at error steps), without positive-class weights on the step heads and without truncating trajectories, for three epochs (6{,}552 steps of eight episodes) with AdamW at learning rate $3\times10^{-4}$; the checkpoint with the highest validation M2 AUPRC is kept. We verified causality on the trained model: changing any action after step $t$ changes the outputs at steps $\le t$ by at most $2.9\times10^{-6}$ (floating-point noise).

\paragraph{CUSUM.} A dedicated change-point detector without learning: a task-conditioned action-bigram model $P(a_t \mid a_{t-1})$ with add-$k$ smoothing and unigram back-off is fitted on the correct training episodes; the per-step surprise $s_t=-\log P(a_t\mid a_{t-1})$ drives the CUSUM statistic $S_t=\max(0, S_{t-1}+s_t-k)$, where $k$ is the mean surprise on the training data, and $p_t=S_t/(S_t+k)$. CUSUM makes no type prediction, so its M4 is 0.

\subsection{Symbolic Baselines}
\label{app:symbolic}
Both symbolic baselines replay the observed actions in the PDDL simulator and therefore use the domain model, which the learned monitors do not receive.

\paragraph{Precondition checker.} An alarm ($p_t=1$) is raised at any step whose action violates its PDDL preconditions in the replayed state; all other steps receive $p_t=0$. Because seven of the nine categories pass strict replay by construction (Section~\ref{app:validation}), it can fire only on the semantic precondition relaxed for substitutions, and it raises an alarm on fewer than 5\% of errorful test episodes.

\paragraph{Re-planner.} The re-planner additionally receives the task's \emph{template-level} formal goal---the subgoals of the task template with object placeholders, e.g.\ ``some bread is toasted and buttered and lies on some plate''---but not the episode's grounded goal, so it does not know which objects the person will use or the correct final state. It uses the generator's per-subgoal STRIPS planner. At every (re-)planning point it scores the admissible bindings of the template placeholders by the number of subgoals already satisfied and by the objects the person has touched, plans for the three best bindings, and keeps the shortest plan. It then tracks the plan: an action that matches the next planned action advances it ($p_t=0$); an action that violates its preconditions raises $p_t=1$; any other action triggers re-planning from the new state, and if the new plan is $\Delta>0$ steps longer than the remaining old plan minus one, $p_t=\Delta/(\Delta+1)$, otherwise $p_t=0$. At the task-end event, $p_t=1$ if no admissible binding satisfies all subgoals. When the person walks to a location outside the planner's task-relevant set, that location is added to the planner's action space before re-planning. Each episode has a 30-minute budget; 60 test episodes (42 errorful, 18 correct) did not finish and are excluded from M1 and M3, as for any method with missing predictions.

\paragraph{Analysis.} The re-planner's M1 (0.887) comes almost entirely from the end-of-task check: this check alone gives M1 0.802, and with the task-end step removed M1 drops to 0.669 $[0.622, 0.714]$. On the steps before task end its AUPRC is 0.118, it raises the first alarm early in 84\% of detected errorful episodes, and it alarms on 62\% of correct episodes: correct executions are not shortest plans, so deviation from a plan is not an error. Its weakest categories are those that leave the goal satisfied, such as spatial misreach and action addition. For reference, the row marked $^\ast$ in Tables~\ref{tab:ci}--\ref{tab:m3-full} gives a privileged variant that receives the episode's grounded goal, i.e.\ the correct final state (33 test episodes did not finish); it is an upper reference for the re-planner and not a monitor that could be deployed.

\subsection{Additional Results}
\label{app:additional}
Table~\ref{tab:ci} gives 95\% confidence intervals for the main results (trajectory-level bootstrap stratified by errorful and correct episodes, 1{,}000 resamples), Table~\ref{tab:complexity} breaks M1 and M2-mid down by template complexity, and Table~\ref{tab:m3-full} gives the full first-alarm distribution. The learned monitors degrade with task complexity (SGT: M1 0.844 on simple, 0.573 on complex templates), whereas the re-planner does not, because its end-of-task goal check does not depend on task length.

\begin{table}[t]
\centering
\small
\setlength{\tabcolsep}{4pt}
\caption{Main results with 95\% confidence intervals (500-episode test set). M2-end: task-end step (omissions); M2-mid: steps before task end (all other errors). Random: M1 0.500, M2 0.036, M2-end 0.274, M2-mid 0.023. $^\ast$Privileged: receives the episode's grounded goal.}
\label{tab:ci}
\begin{tabular}{lcccc}
\toprule
Method & M1 & M2 & M2-end & M2-mid \\
\midrule
Seq2Seq & 0.496 [0.44, 0.55] & 0.376 [0.33, 0.43] & 0.656 [0.58, 0.73] & 0.132 [0.11, 0.17] \\
Seq2Seq+SG & 0.554 [0.50, 0.60] & 0.394 [0.34, 0.45] & 0.665 [0.58, 0.75] & 0.151 [0.13, 0.18] \\
SGT & 0.778 [0.73, 0.82] & 0.656 [0.61, 0.70] & 0.791 [0.73, 0.85] & 0.553 [0.49, 0.62] \\
CUSUM & 0.645 [0.59, 0.69] & 0.039 [0.04, 0.04] & 0.235 [0.20, 0.28] & 0.029 [0.03, 0.03] \\
\midrule
Precondition check & 0.513 [0.50, 0.52] & 0.059 [0.05, 0.08] & 0.274 [0.24, 0.31] & 0.062 [0.04, 0.09] \\
Re-planner & 0.887 [0.86, 0.91] & 0.379 [0.32, 0.45] & 0.690 [0.63, 0.76] & 0.118 [0.07, 0.17] \\
Re-planner (episode goal)$^\ast$ & 0.922 [0.90, 0.94] & 0.371 [0.31, 0.44] & 0.673 [0.61, 0.74] & 0.125 [0.08, 0.18] \\
\bottomrule
\end{tabular}
\end{table}

\begin{table}[t]
\centering
\small
\setlength{\tabcolsep}{5pt}
\caption{M1 and M2-mid by template complexity on the test set (121 simple, 291 medium, 88 complex episodes).}
\label{tab:complexity}
\begin{tabular}{lcccccc}
\toprule
 & \multicolumn{2}{c}{Simple} & \multicolumn{2}{c}{Medium} & \multicolumn{2}{c}{Complex} \\
\cmidrule(lr){2-3}\cmidrule(lr){4-5}\cmidrule(lr){6-7}
Method & M1 & M2-mid & M1 & M2-mid & M1 & M2-mid \\
\midrule
Seq2Seq & 0.614 & 0.189 & 0.498 & 0.140 & 0.349 & 0.047 \\
Seq2Seq+SG & 0.706 & 0.229 & 0.536 & 0.152 & 0.391 & 0.154 \\
SGT & 0.844 & 0.690 & 0.797 & 0.546 & 0.573 & 0.406 \\
CUSUM & 0.666 & 0.178 & 0.654 & 0.039 & 0.633 & 0.031 \\
\midrule
Precondition check & 0.507 & 0.055 & 0.516 & 0.076 & 0.509 & 0.039 \\
Re-planner & 0.866 & 0.073 & 0.897 & 0.140 & 0.894 & 0.150 \\
Re-planner (episode goal)$^\ast$ & 0.900 & 0.098 & 0.940 & 0.180 & 0.923 & 0.063 \\
\bottomrule
\end{tabular}
\end{table}

\begin{table}[t]
\centering
\small
\setlength{\tabcolsep}{4pt}
\caption{M3 first-alarm distribution at $\theta{=}0.5$ on the test set. Missed, early ($\Delta<0$), on target ($\Delta=0$), and late ($\Delta>0$) are shares of the 175 episodes whose first error occurs before task end; false alarm is the share of correct episodes with an alarm. Omissions: share of the 137 omission episodes alarmed at the task-end step / before it. Episodes on which the re-planner timed out are excluded from its rows (145 / 125 / 170 episodes for the template-goal re-planner and 155 / 132 / 180 for the privileged one).}
\label{tab:m3-full}
\begin{tabular}{lcccccc}
\toprule
Method & Missed & Early & On target & Late & False alarm & Omission: end / before \\
\midrule
Seq2Seq & 0.99 & 0.00 & 0.00 & 0.01 & 0.10 & 0.46 / 0.00 \\
Seq2Seq+SG & 0.98 & 0.00 & 0.00 & 0.02 & 0.05 & 0.38 / 0.00 \\
SGT & 0.52 & 0.04 & 0.38 & 0.06 & 0.12 & 0.58 / 0.01 \\
CUSUM & 0.06 & 0.77 & 0.07 & 0.10 & 0.81 & 0.01 / 0.80 \\
\midrule
Precondition check & 0.95 & 0.00 & 0.05 & 0.00 & 0.00 & 0.00 / 0.00 \\
Re-planner & 0.10 & 0.75 & 0.05 & 0.10 & 0.62 & 0.33 / 0.66 \\
Re-planner (episode goal)$^\ast$ & 0.13 & 0.74 & 0.05 & 0.08 & 0.33 & 0.59 / 0.41 \\
\bottomrule
\end{tabular}
\end{table}

\subsection{LLM Baselines}
\label{app:llm}

\paragraph{Protocol.} LLMs are evaluated prefix by prefix with the same input as every other monitor. For each evaluated step $t$ of an episode, one request contains the system prompt below and a user message with the task name, the initial scene (every object with its class, location, and initial states), and the actions $a_{0:t}$; the model returns a JSON object for step $t$ only, with \texttt{p\_instant} ($p_t$) and nine cumulative category probabilities (\texttt{p\_types}). No future action, trajectory length, goal condition, reference execution, or label is included; the task-end event appears only at the final step. On the 60-episode subset this gives 1{,}201 requests per configuration. Before running, we checked all 1{,}201 user prompts automatically for these properties.

\paragraph{API settings.} All calls use OpenRouter with temperature 0. Because the default reasoning behaviour of the same model differs across providers (e.g.\ DeepSeek V4 Flash reasons by default on one provider and not on another), each configuration is pinned to a single provider with fallback disabled, and reasoning is set explicitly: disabled, an effort level, or a token budget (Table~\ref{tab:llm-suite}). Requests time out after 120--600\,s and are retried twice; steps whose response cannot be parsed are re-requested once with a doubled output limit. All thirteen configurations completed every step. Every response is logged verbatim, together with the reasoning text when the provider returns it, token counts, cost, and the serving provider; the logs are released with the benchmark. The total cost of the thirteen runs was US\$107.

\paragraph{System prompt.} Reproduced verbatim (line breaks added for display; an em dash is typeset as ``--'').
{\small
\begin{verbatim}
You are monitoring a human performing a household task in real time, ONE STEP AT A
  TIME.
You see the task name, the initial scene, and the actions from step 0 up to and
  including the current step.
You CANNOT see any future action, and you are NOT told how long the task will
  last.
The last step of every episode is the event `human-task-end h`: the person
  declares the WHOLE task finished.
It appears only when it happens. At that step you can judge whether some step the
  task needs was never done.

For the CURRENT step, output two judgments:
- p_instant (0.0-1.0): probability that an error occurs AT the current step (the
  current action is where an error happens).
- p_types: for each of the 9 error categories, the probability that an error of
  that category has occurred
  at or before the current step (cumulative over the episode so far).

IMPORTANT GUIDANCE:
- You are NOT given the formal goal conditions, only the task name. Infer what the
  task requires from the task name and the scene, as a person watching would.
- By default assume the current step is correct (p_instant near 0). Give p_instant
  > 0.5 only with clear evidence that an error happens AT this step.
- A step is correct if it is a reasonable action toward the task, even if it is
  roundabout or inefficient. Harmless redundancy is NOT an error: walking back and
  forth between places the task uses, picking something up and putting it back, or
  repeating an action immediately when the repetition has no additional effect.
- p_instant marks only the step where an error happens, not later steps that are
  merely its consequences.
- p_types is cumulative: once an error of some category has happened, keep that
  category high at all later steps, including the final task-end step.

Error categories (the step in brackets is where p_instant should be high):
- OMISSION: a step the task requires was never performed. This can only be
  confirmed when the person declares the task finished. [the `human-task-end`
  step]
- SEQUENCE: two steps that must happen in a specific order were done in the wrong
  order, e.g. applying butter to bread before toasting it. [the step done out of
  order]
- ANTICIPATION: an item is stored, served or packed before it receives the
  processing the task requires. [the step that commits the unprocessed item:
  putting it into its final place, closing the container that holds it, or
  delivering that container to its final place]
- PERSEVERATION: after doing something else in between, the person repeats an
  action that should be done only once, e.g. adds sugar to the coffee, butters the
  toast, then adds sugar to the coffee again. [the repeated action]
- OBJECT_SUBSTITUTION: a similar object of the same kind is used in place of the
  object the task needs. [the first step that uses the substitute]
- TOOL_SUBSTITUTION: a similar tool of the same kind is used in place of the
  appropriate tool. [the first step that uses the substitute tool]
- ACTION_ADDITION: an item unrelated to the task is put into or onto the task's
  target container, plate or tray. Removing it later does not undo the error. [the
  step that puts it there]
- CAPTURE_ERROR: an action the task needs is applied to an unrelated object of a
  different kind instead of the task's object. [the step that acts on the
  unrelated object]
- SPATIAL_MISREACH: the person goes to a storage location that has nothing to do
  with the task, opens it, and leaves without taking anything. [the step of
  leaving]

OUTPUT FORMAT -- respond with ONLY valid JSON (no markdown, no commentary):
{"step": <int>, "p_instant": <float>, "p_types": {"OMISSION": <float>, "SEQUENCE":
  <float>, "ANTICIPATION": <float>, "PERSEVERATION": <float>,
  "OBJECT_SUBSTITUTION": <float>, "TOOL_SUBSTITUTION": <float>, "ACTION_ADDITION":
  <float>, "CAPTURE_ERROR": <float>, "SPATIAL_MISREACH": <float>}}
Output exactly ONE such object, for the current step only.
\end{verbatim}
}

\noindent\begin{minipage}{\linewidth}
\paragraph{User prompt template.}\leavevmode\par
{\small
\begin{verbatim}
Task: <task name, underscores replaced by spaces>

Initial scene: <object>:<CLASS>@<location>[<states>] ...

Actions observed so far:
Step 0: <action> h <arg1> <arg2> ...
...
Step t: <action> h <arg1> ...

Judge the current step (Step t). Output the JSON object for step t only.
\end{verbatim}
}
\end{minipage}

\subsection{Full LLM Results}
\label{app:full-results}

Table~\ref{tab:llm-results} reports all thirteen configurations on the 60-episode subset (30 errorful, 30 correct; 1{,}201 evaluated steps, 33 of them error steps; M2 random 0.027, M2-end 0.067, M2-mid 0.025), with 95\% bootstrap intervals, together with the other baselines re-scored on the same episodes; Table~\ref{tab:llm-m3} gives the first-alarm distribution.

\paragraph{Key observations.}
\begin{enumerate}
\item \textbf{Models cannot be ranked on M1.} With 60 episodes, the M1 intervals of all configurations overlap. The best upper bounds stay below the human reference on both M1 (0.837 vs.\ 0.967) and M2 (0.482 vs.\ 0.748).
\item \textbf{Reasoning matters for localization and diagnosis.} With reasoning enabled, the four larger models reach M2-mid 0.18--0.48 and M4 0.43--0.52; with reasoning disabled, M2-mid is at most 0.15 and M4 at most 0.31, and Qwen3.6 and DeepSeek V4 Flash output near-zero scores at almost every step. At the steps before task end, the best reasoning configurations (DeepSeek V4 Pro 0.478, Claude Sonnet 4.6 0.451) are on par with SGT on the same episodes (0.455); none of them confirms omissions at task end well (M2-end at most 0.22).
\item \textbf{More effort does not help further.} GPT-5.5 low and medium are indistinguishable on every metric (M2 0.238 vs.\ 0.231), at 65\% higher cost for medium.
\item \textbf{Reasoning trades misses for false alarms.} The reasoning configurations give the most on-target first alarms (42--50\% of errorful episodes) but alarm on 30--77\% of correct episodes; without reasoning, most models rarely alarm at all.
\end{enumerate}

\begin{table}[H]
  \centering
  \small
  \caption{LLM configurations. Each configuration is pinned to one OpenRouter provider (fallback disabled); reasoning is set explicitly. $^a$Reports one reasoning token per request with reasoning disabled, and no reasoning text. $^b$The provider rejects requests with reasoning disabled (``reasoning is mandatory''), so the lowest effort is used.}
  \label{tab:llm-suite}
  \footnotesize
  \setlength{\tabcolsep}{3pt}
  \begin{tabular}{l l l l r}
  \toprule
  Model & OpenRouter ID & Provider & Reasoning & Max tokens \\
  \midrule
  Qwen3.6-35B-A3B & \texttt{qwen/qwen3.6-35b-a3b} & DeepInfra (fp8) & disabled & 512 \\
  Gemini 3.1 Flash Lite & \texttt{google/gemini-3.1-flash-lite} & Google AI Studio & disabled & 512 \\
  Kimi K2.6 & \texttt{moonshotai/kimi-k2.6} & Moonshot AI & disabled$^a$ & 512 \\
  Grok 4.3 & \texttt{x-ai/grok-4.3} & xAI & disabled & 512 \\
  GLM-5.1 & \texttt{z-ai/glm-5.1} & Z.AI (fp8) & disabled & 512 \\
  DeepSeek V4 Flash & \texttt{deepseek/deepseek-v4-flash} & Novita (fp8) & disabled & 512 \\
  GPT-5.4 mini & \texttt{openai/gpt-5.4-mini} & OpenAI & disabled & 512 \\
  MiniMax-M2.7 & \texttt{minimax/minimax-m2.7} & MiniMax & low$^b$ & 4{,}000 \\
  DeepSeek V4 Pro & \texttt{deepseek/deepseek-v4-pro} & Alibaba (fp8) & medium & 16{,}000 \\
  Gemini 3.1 Pro & \texttt{google/gemini-3.1-pro-preview} & Google AI Studio & medium & 8{,}000 \\
  Claude Sonnet 4.6 & \texttt{anthropic/claude-sonnet-4.6} & Anthropic & 2{,}000 tokens & 4{,}000 \\
  GPT-5.5 & \texttt{openai/gpt-5.5} & OpenAI & low & 8{,}000 \\
  GPT-5.5 & \texttt{openai/gpt-5.5} & OpenAI & medium & 8{,}000 \\
  \bottomrule
  \end{tabular}
\end{table}

\begin{table}[H]
\centering
\small
\setlength{\tabcolsep}{3pt}
\caption{LLM results on the 60-episode subset, sorted by M1, with the other baselines re-scored on the same episodes. M4 includes correct episodes (Section~\ref{app:supervised}). 95\% intervals from a trajectory-level bootstrap stratified by errorful and correct episodes (1{,}000 resamples).}
\label{tab:llm-results}
\begin{tabular}{lccccc}
\toprule
Model & M1 & M2 & M2-end & M2-mid & M4 \\
\midrule
Kimi K2.6 & 0.718 [0.60, 0.84] & 0.140 [0.06, 0.25] & 0.067 & 0.154 & 0.179 \\
Claude Sonnet 4.6 (2{,}000 tokens) & 0.697 [0.56, 0.82] & 0.325 [0.20, 0.48] & 0.111 & 0.451 & 0.432 \\
DeepSeek V4 Pro (medium) & 0.679 [0.55, 0.81] & 0.286 [0.17, 0.44] & 0.218 & 0.478 & 0.480 \\
GPT-5.5 (low) & 0.631 [0.48, 0.77] & 0.238 [0.14, 0.38] & 0.133 & 0.394 & 0.489 \\
GPT-5.5 (medium) & 0.617 [0.47, 0.75] & 0.231 [0.15, 0.37] & 0.116 & 0.430 & 0.477 \\
Grok 4.3 & 0.584 [0.48, 0.69] & 0.086 [0.03, 0.18] & 0.067 & 0.093 & 0.107 \\
GPT-5.4 mini & 0.579 [0.45, 0.71] & 0.055 [0.03, 0.12] & 0.067 & 0.059 & 0.143 \\
GLM-5.1 & 0.576 [0.45, 0.72] & 0.077 [0.05, 0.13] & 0.081 & 0.114 & 0.305 \\
Gemini 3.1 Pro (medium) & 0.572 [0.44, 0.69] & 0.119 [0.07, 0.20] & 0.069 & 0.175 & 0.519 \\
Gemini 3.1 Flash Lite & 0.564 [0.49, 0.65] & 0.061 [0.03, 0.16] & 0.067 & 0.064 & 0.086 \\
MiniMax-M2.7 (low) & 0.527 [0.38, 0.67] & 0.079 [0.04, 0.17] & 0.113 & 0.090 & 0.192 \\
Qwen3.6-35B-A3B & 0.517 [0.50, 0.55] & 0.057 [0.02, 0.12] & 0.067 & 0.059 & 0.109 \\
DeepSeek V4 Flash & 0.500 [0.50, 0.50] & 0.027 [0.02, 0.03] & 0.067 & 0.025 & 0.099 \\
\midrule
Seq2Seq & 0.336 [0.20, 0.48] & 0.107 [0.05, 0.21] & 0.375 & 0.085 & 0.452 \\
Seq2Seq+SG & 0.350 [0.20, 0.50] & 0.114 [0.07, 0.22] & 0.277 & 0.112 & 0.533 \\
SGT & 0.683 [0.53, 0.80] & 0.423 [0.29, 0.59] & 0.458 & 0.455 & 0.490 \\
CUSUM & 0.692 [0.56, 0.82] & 0.036 [0.03, 0.05] & 0.119 & 0.035 & 0.000 \\
Precondition check & 0.517 [0.50, 0.55] & 0.057 [0.02, 0.12] & 0.067 & 0.059 & 0.000 \\
Re-planner & 0.871 [0.76, 0.96] & 0.113 [0.05, 0.26] & 0.273 & 0.158 & 0.084 \\
\bottomrule
\end{tabular}
\end{table}

\begin{table}[H]
\centering
\small
\setlength{\tabcolsep}{5pt}
\caption{M3 first-alarm distribution at $\theta{=}0.5$ on the 60-episode subset: shares of the 26 errorful episodes whose first error occurs before task end (missed, early, on target, late) and false-alarm rate on the 30 correct episodes; for the re-planner, 22 and 28 episodes, excluding timeouts.}
\label{tab:llm-m3}
\begin{tabular}{lccccc}
\toprule
Model & Missed & Early & On target & Late & False alarm \\
\midrule
Kimi K2.6 & 0.46 & 0.38 & 0.12 & 0.04 & 0.27 \\
Claude Sonnet 4.6 (2{,}000 tokens) & 0.31 & 0.08 & 0.50 & 0.12 & 0.30 \\
DeepSeek V4 Pro (medium) & 0.27 & 0.19 & 0.42 & 0.12 & 0.47 \\
GPT-5.5 (low) & 0.15 & 0.08 & 0.46 & 0.31 & 0.57 \\
GPT-5.5 (medium) & 0.23 & 0.12 & 0.46 & 0.19 & 0.50 \\
Grok 4.3 & 0.69 & 0.27 & 0.04 & 0.00 & 0.13 \\
GPT-5.4 mini & 0.73 & 0.19 & 0.04 & 0.04 & 0.07 \\
GLM-5.1 & 0.19 & 0.27 & 0.15 & 0.38 & 0.67 \\
Gemini 3.1 Pro (medium) & 0.08 & 0.23 & 0.42 & 0.27 & 0.77 \\
Gemini 3.1 Flash Lite & 0.81 & 0.08 & 0.08 & 0.04 & 0.03 \\
MiniMax-M2.7 (low) & 0.54 & 0.23 & 0.15 & 0.08 & 0.47 \\
Qwen3.6-35B-A3B & 0.96 & 0.00 & 0.04 & 0.00 & 0.00 \\
DeepSeek V4 Flash & 1.00 & 0.00 & 0.00 & 0.00 & 0.00 \\
\midrule
Seq2Seq & 1.00 & 0.00 & 0.00 & 0.00 & 0.07 \\
Seq2Seq+SG & 1.00 & 0.00 & 0.00 & 0.00 & 0.07 \\
SGT & 0.73 & 0.04 & 0.23 & 0.00 & 0.07 \\
CUSUM & 0.04 & 0.85 & 0.04 & 0.08 & 0.87 \\
Precondition check & 0.96 & 0.00 & 0.04 & 0.00 & 0.00 \\
Re-planner & 0.09 & 0.82 & 0.00 & 0.09 & 0.64 \\
\bottomrule
\end{tabular}
\end{table}

\subsection{Expert Review}
\label{app:expert}
To obtain an initial check of generator fidelity, two domain experts (neuropsychology; nursing care) jointly reviewed 24 randomly selected NAT-aligned episodes (12 errorful, 12 correct) without access to the labels. They judged 10 of the 12 injected errors plausible, identified 11 of the 12 correct episodes as correct, and matched the generator's error category on 8 of the 12 errorful episodes. The experts noted that the NAT alignment yields realistic tasks, whereas the plausibility of an error depends on the scene, which is consistent with the scene-specific feasibility rules of Section~\ref{app:generation}. Mapping the generated errors to NAT error instances gives three direct and four structural matches. These results provide initial expert support for the generator, but they are not a patient-level validation; results of further expert validation will be released on the project website.

\end{document}